\documentclass{article} 
\usepackage{iclr2027_conference,times}

\usepackage{amsmath,amsfonts,bm}

\def\eqref#1{equation~\ref{#1}}

\def\1{\bm{1}}

\DeclareMathAlphabet{\mathsfit}{\encodingdefault}{\sfdefault}{m}{sl}
\SetMathAlphabet{\mathsfit}{bold}{\encodingdefault}{\sfdefault}{bx}{n}

\usepackage{hyperref}
\usepackage{url}
\usepackage{wrapfig}
\usepackage{graphicx}
\usepackage{booktabs}
\usepackage[table]{xcolor}
\definecolor{lightpurple}{RGB}{235,225,250}
\usepackage{subcaption}
\usepackage{multirow}
\usepackage{makecell}
\usepackage{amsmath}
\usepackage{amssymb}
\usepackage{titletoc}
\usepackage{hyperref}

\newcommand{\OurModel}{\textit{CellRefine}}
\newcommand{\SW}[1]{\textcolor{red}{#1}}

\title{Prototype Guided Post-pretraining \\ for Single-Cell Representation Learning}

\author{
Sachini Weerasekara$^{1}$ \quad
Natasha Darras$^{2}$ \quad
Sagar Kamarthi$^{1}$ \quad
Colles Price$^{2}$ \quad
Jacqueline Isaacs$^{1}$ \\[1ex]
$^{1}$Department of Mechanical and Industrial Engineering, Northeastern University,
Boston, MA 02115, USA \\
$^{2}$Takeda Pharmaceuticals, Cambridge, MA 02142, USA \\[1ex]
}

\iclrfinalcopy 

\begin{document}

\maketitle
\fancyhead{}
\vspace{-0.5cm}
\begin{abstract}
Single-cell representation learning (SCRL) from gene expression data offers a way to uncover the complex regulatory logic underlying cellular function. Inspired by large language models in natural language modeling, several single-cell pretrained models have recently been proposed that treat genes as tokens and cells as sentences. However, these models are fundamentally limited by the long-tailed nature of cell-type distributions and struggle to generalize under covariate shifts in gene expression data. While fine-tuning is often used to mitigate these issues, we observe that performance remains bounded. To address this challenge, we introduce \textit{\OurModel}, a post-pretraining method that operates between the pretraining and fine-tuning stages of a single-cell foundation model. \textit{\OurModel} uses a multi-faceted objective that incorporates marker-gene sets as structural priors to guide post-pretraining and refine the latent embedding manifold of cells. Across multiple computational biology tasks, empirical results show that \textit{\OurModel} consistently improves downstream performance, yielding gains up to 15\%.
\end{abstract}

\section{Introduction}

Advances in machine learning are rapidly transforming our ability to study complex cellular systems, providing powerful computational frameworks for decoding high-dimensional single-cell data and enabling breakthroughs in precision diagnostics~\cite{xia2025precise}, personalized therapies~\cite{tan2025prediction}, and proactive disease prevention~\cite{shmatko2025learning}. Among the most promising directions is single-cell representation learning (SCRL), which leverages cell–gene expression data derived from modalities such as single-cell RNA sequencing and spatial transcriptomics to learn rich representations of cell identities and functions. 

Inspired by the success of large language models in natural language modeling, recent work has proposed single-cell pretrained models for SCRL, framing genes as tokens and cells as sentences~\cite{theodoris2023transfer, cui2024scgpt}. These models open new opportunities for uncovering latent cellular states and disentangling complex gene regulatory programs, with applications ranging from reconstructing developmental trajectories and revealing disease mechanisms to predicting drug-induced perturbations for therapeutic discovery.

\begin{wrapfigure}{r}{0.45\columnwidth}
\vspace{-12pt}
  \centering
  \includegraphics[width=0.45\textwidth]{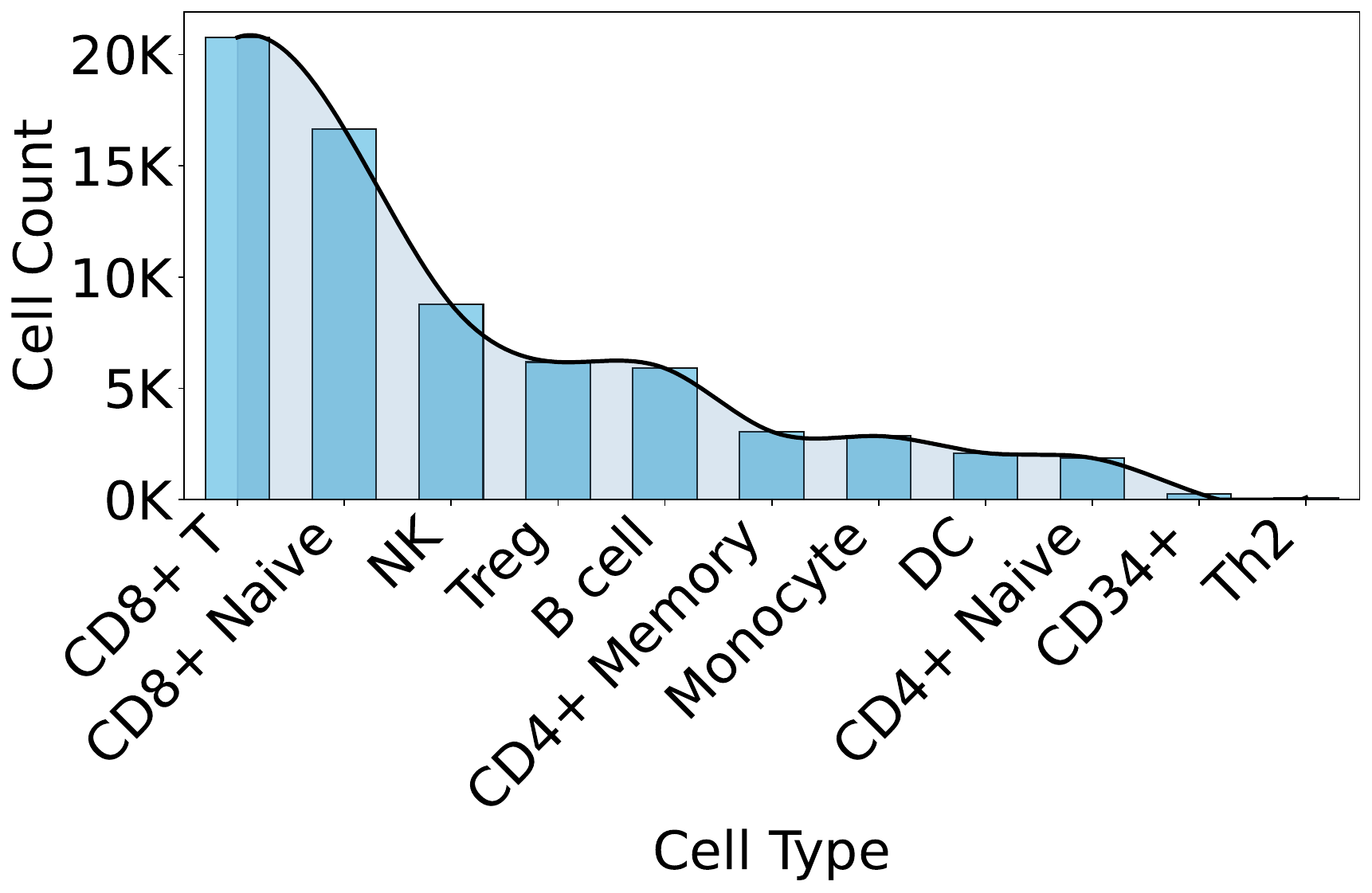}
  \caption{Long-tail distribution of cell types in the blood cell dataset~\cite{zheng2017massively}.}
  \label{fig:long-tail-distribution}
  \vspace{-12pt}
\end{wrapfigure}

Despite this widespread applicability, existing pretrained models for SCRL struggle to learn equally rich representations across the full spectrum of cell types \cite{navidi2025adaptive} and are generally performance-bound. We hypothesize that these limitations stem from two primary factors. First, the long-tailed distributions inherent in single-cell datasets lead to noisy or collapsed embeddings for underrepresented cell populations. For example, Figure~\ref{fig:long-tail-distribution} illustrates the extreme disparity in sample counts between cell types in a standard blood cell dataset~\cite{zheng2017massively}, with further analyses provided in Appendix ~\ref{sec:additional-dataset-details}. Second, pervasive covariate shifts in gene expression measurements create significant distribution gaps between training and inference data. These technical changes often mask true biological signals, forcing models to overfit batch-specific patterns, challenging robust downstream generalization.

The extreme imbalance between cell types reflects labeling challenges, technical artifacts, and rare biology, with critical but rare populations, such as disease-initiating stem cells or specialized immune cells, often constitute less than 0.1\% of a sample \cite{azizi2018single, grun2015single, villani2017single}. Our analysis confirms this long-tail pattern across major public benchmarks, revealing power-law tail exponents as low as $\alpha \approx 0.2{-}0.6$ (see Appendix\SW{~\ref{long-tail-analysis}}). Such a severe skew biases pretrained models toward abundant cell types, causing the latent space to overlook rare but clinically significant cell populations.

Even when underrepresented cells are captured, their biological signal is frequently buried under pervasive covariate shifts. Driven by technical artifacts like batch effects and varying sequencing depths, these shifts alter the marginal distribution of gene expression levels without changing the underlying cell-type identity. Because most single-cell pretrained models are sensitive to these high-dimensional fluctuations, they often fail to generalize across different sequencing platforms or experimental protocols. Although fine-tuning on individual datasets is commonly used to address these issues, its effectiveness is often limited because it begins from an unstable latent embedding manifold in the foundation model, restricting its ability to learn more refined cell representations.

To address these limitations, we introduce \textit{{\OurModel}}, a post-pretraining framework for single-cell foundation models that is designed to better capture rare cell types in the long tail of the cell-type distribution while improving robustness to covariate shift. Rather than relying on fine-tuning alone from an unstable latent embedding manifold, \text{{\OurModel}} injects biologically grounded inductive biases during post-pretraining to reshape the latent embedding space into a more stable and discriminative structure. Specifically, \text{{\OurModel}} leverages mechanistic biological priors derived from known marker gene programs to guide the model toward distinguishing rare cell types whose signals are often overlooked or confounded by technical noise. In addition, to separate closely related cell types within the same parent lineage in the cell ontology, \text{{\OurModel}} incorporates a lineage-aware regularization loss that explicitly encourages the separation of cells sharing the same parent lineage.

Together, the marker gene program guidance and cell ontology-aware regularization provide stronger biological grounding and help the model learn more discriminative, stable, and generalizable representations, thereby mitigating both long-tail and covariate-shift challenges. By explicitly incorporating prior knowledge about cell identity and lineage structure, these objectives encourage the learned representation space to better reflect biologically meaningful relationships rather than relying solely on statistical regularities in the pretraining data. More broadly, this work asks the following question: \textbf{To what extent can biologically informed inductive biases and explicit learning guidance improve the downstream performance, robustness, and transferability of single-cell foundation models?}

To assess the utility of {\OurModel}, we conduct extensive experiments across three core tasks in computational biology: 1) cell identity prediction 2) spatial transcriptomic imputation and 3) gene perturbation response prediction across 11 datasets spanning blood, pancreas, myeloid, liver, brain, lung, and heart tissues. Empirical evaluation demonstrates that {\OurModel} consistently surpasses state-of-the-art single-cell pretrained models across all three tasks and organ types, yielding up to 15\% performance improvements.

\begin{figure}[h]
\vspace{-12pt}
  \centering
  \includegraphics[width=1\linewidth]{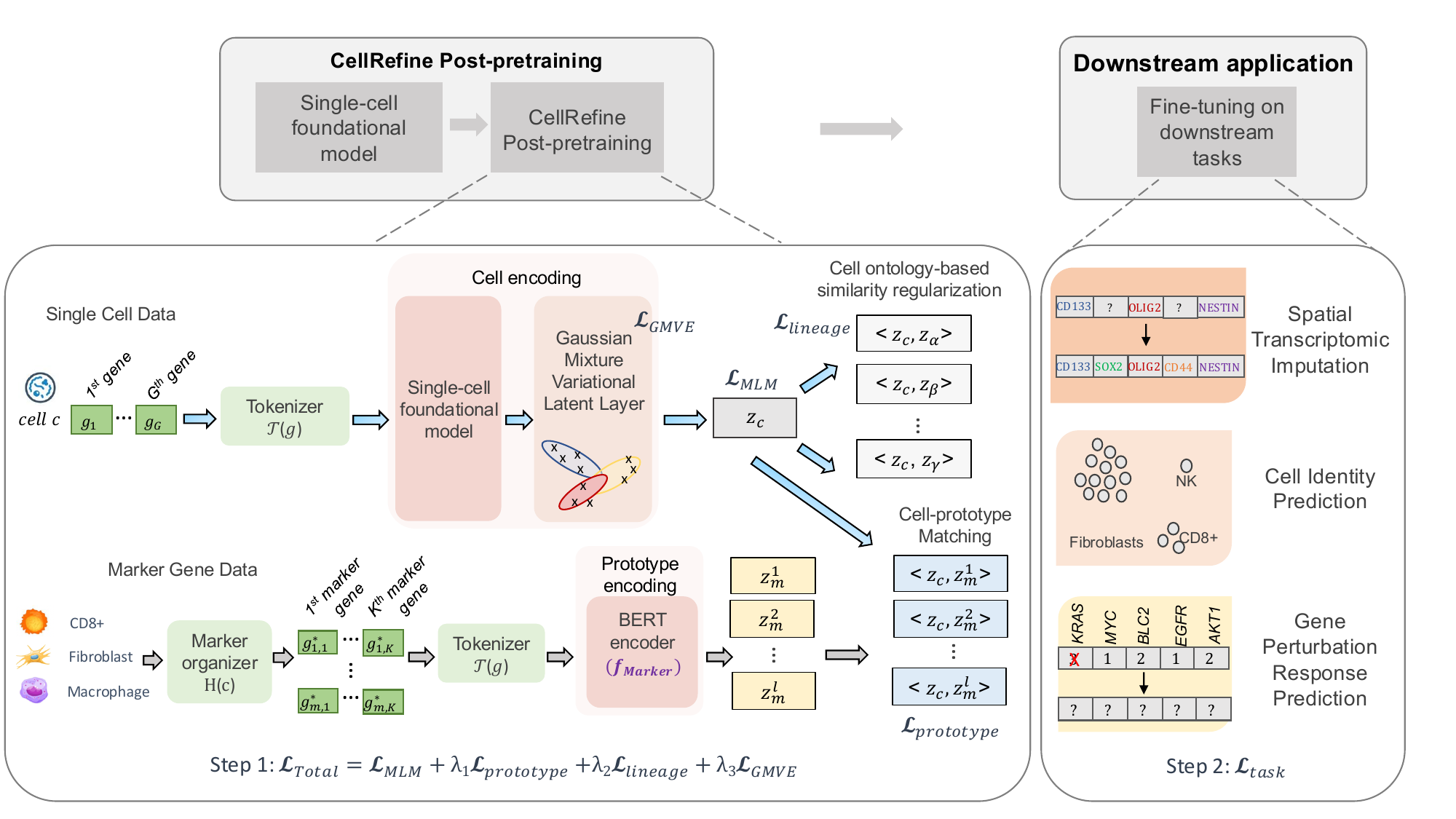}
    \caption{Overview of \text{\OurModel}, a post-pretraining method for single-cell foundation models. \text{\OurModel} refines the latent cell embedding space of a given foundation model on a target cell dataset before task-specific fine-tuning.}
  \label{fig:overall-method}
\vspace{-12pt}
\end{figure}

\section{Methodology}

In this section, we elaborate on the details of our method behind {\OurModel}. Additionally, an overview of the proposed method as illustrated in Figure~\ref{fig:overall-method}.

\subsection{Post-pretaining with {\OurModel}}

{\OurModel} is an intermediate post-pretraining technique designed to better align a single-cell pretrained model embedding manifold and prepare it for downstream fine-tuning. Specifically, {\OurModel} is applied after pretraining and before downstream fine-tuning. Such intermediate adaptation strategies have recently been proposed in computer vision as an effective way to refine pretrained models for specific data characteristics and downstream objectives~\cite{yamaguchi2025post}.

Specifically, let \(f_{\theta^{\mathrm{pt}}}\) denote a single-cell pretrained model after the initial pretraining stage, where pretrained \(\theta^{\mathrm{pt}}\) are often obtained by optimizing a language modeling objective (refer to Appendix~\ref{scrl-token-prediction} for the details of pretraining objective formulation). We then apply \text{\OurModel} as an intermediate post-pretraining stage to adapt this pretrained model towards a given dataset, yielding an updated model \(f_{\theta^{\mathrm{pp}}}\), where \(\theta^{\mathrm{pp}}\) is initialized from \(\theta^{\mathrm{pt}}\) and optimized under our post-pretraining method. Finally, for a downstream task of interest, we fine-tune the post-pretrained model to obtain \(f_{\theta^{\mathrm{ft}}}\), where \(\theta^{\mathrm{ft}}\) is initialized from \(\theta^{\mathrm{pp}}\) and optimized using the task-specific objective. Thus, the overall training setup can be written as,
\vspace{-0.1cm}
\[
f_{\theta^{\mathrm{pt}}}
\;\xrightarrow{\text{\OurModel}}\;
f_{\theta^{\mathrm{pp}}}
\;\xrightarrow{\text{Downstream Fine-tuning}}\;
f_{\theta^{\mathrm{ft}}}
\]

\subsection{Prototype-guided Learning}

The {\OurModel} post-pretraining stage is driven by our novel \textit{prototype-guided Learning} mechanism, a training method that grounds a given pretrained model on mechanistic biological priors to guide the learning towards refined cell representations overall. To this end, we introduce two biological priors to the learning process: 1) gene marker program-based guided masked language modeling and 2) cell similarity regularization based on cell ontology. 

\subsubsection{Gene marker program-based guided masked language modeling}

Gene markers are specific genes whose expression patterns are characteristic of particular cell types, effectively serving as biological “labels” that distinguish one cell from another. In single-cell analysis, researchers often rely on these well-established markers to manually identify and classify cells, interpreting clusters or assigning cell types based on known marker genes~\cite{dumitrascu2021optimal}. This suggests that incorporating marker genes during the post-pretraining stage can provide a biologically meaningful supervisory signal, helping to further structure the pretrained embedding space.

However, it is important to note that the full set of marker genes for a given cell type is typically unknown; only a subset of markers is often identified. Consequently, marker genes provide a partial learning signal rather than a complete one. Nevertheless, known markers capture key functional and identity-related features of cells, making them a useful source of guidance. In {\OurModel}, we tap on this unexplored opportunity to further refine the embeddings of a pretrained model.

To start, we construct marker gene sets for each cell type. For this, we integrate multiple publicly available marker gene resources as described in Appendix~\ref{marker construction}. In short, for each cell type \(c_i\) where \(i \in \{1, \dots, l\}\) where $l$ is the number of cell types in a given dataset, we identify \(K\) marker genes and order them based on their position in the cell ontology using a marker organizer function \(H(c_i)\) (refer to Appendix~\ref{marker construction} for the implementation details of \(H(c_i)\)). We refer to this ordered sequence of marker genes as a \textit{prototype} of the cell type. Formally, the output of \(H(c_i)\) is an ordered sequence of marker genes $(g^*_{i,1}, \dots, g^*_{i,K})$ for the cell type $c_i$.

These ordered marker genes are tokenized using the tokenizer \(\tau(\cdot)\) and then encoded with a learnable model \(f_{\text{marker}}(\cdot)\). We call this process \textit{prototype encoding}, which produces an embedding for each cell-type prototype and provides a biologically informed reference in the latent space.

\vspace{-0.1cm}
\begin{equation}
z^i_m = f_{\text{marker}}(\tau(g^*_{i,1}, \dots, g^*_{i,K}))
\end{equation}

We perform this procedure for each cell type in a given dataset to get all prototype embeddings \(\{z^1_m, \dots, z^l_m\}\). Then we define a \textit{prototype-guided regularization loss} \(\mathcal{L}_{c \rightleftarrows p}\) to use them for guided post-pretraining. Specifically, let \(z_c^i\) denote the embedding of cell \(c_i\) obtained from a given single-cell pretrained model. We compute the cosine similarity between the cell embedding \(z_c^i\) and each prototype embedding \(z_m^j\).
\vspace{-0.1cm}
\begin{equation}
s_{i,j} = \cos(z_c^i, z_m^j) = \frac{z_c^i \cdot z_m^j}{\|z_c^i\| \, \|z_m^j\|}
\end{equation}

We then define a cross-entropy loss using these similarities, treating the true prototype of the cell as the correct label. 
\vspace{-0.1cm}
\begin{equation}
\mathcal{L}_{prototype} = - \sum_{i=1}^m \log \frac{\exp(s_{i,i})}{\sum_{j=1}^m \exp(s_{i,j})}
\end{equation}

This regularization encourages each cell embedding \(z_c^i\) to align closely with its corresponding prototype \(z_m^i\) than with other prototypes, injecting biologically meaningful structure into the representation space. As illustrated in Figure~\ref{fig:overall-method}, the ground-truth cell-type label selects the prototype, while its marker genes define the prototype structure; the prototype loss is therefore a supervised objective.

\subsubsection{Cell similarity regularization based on cell ontology}

While prototype-guided training helps the model capture inter-cell-type differences, cell types that share the same parent lineage in the cell ontology may still remain insufficiently separated in the embedding space. This is particularly challenging for closely related cell types, whose transcriptional programs often exhibit substantial overlap. To further refine these local structures, \text{{\OurModel}} incorporates a lineage-aware regularization loss that explicitly encourages the separation of cells that share the same parent lineage and ontology level.

Specifically, let \(\mathcal{R}\) denote the set of cell-type pairs \((c_i, c_j)\) such that \(c_i\) and \(c_j\) share the same parent lineage and belong to the same ontology level. During training, for each batch, we identify all pairs in \(\mathcal{R}\) that are present in the batch and compute the cosine similarity between their cell embeddings,
\vspace{-0.1cm}
\begin{equation}
r_{i,j} = \cos(z_c^i, z_c^j) = \frac{z_c^i \cdot z_c^j}{\|z_c^i\| \, \|z_c^j\|}
\end{equation}

Because these cell types are semantically close in the ontology yet represent distinct biological identities, we seek to prevent their embeddings from collapsing together. We therefore define a lineage-aware separation loss that penalizes excessive similarity among such related cell-type pairs.
\vspace{-0.1cm}
\begin{equation}
\mathcal{L}_{\mathrm{lineage}} = \frac{1}{|\mathcal{R}_b|} \sum_{(i,j) \in \mathcal{R}_b} \max(0, r_{i,j}-m)
\end{equation}
where $m$ is a separation margin and \(\mathcal{R}_b \subseteq \mathcal{R}\) denotes the subset of valid cell-type pairs present in the current training batch.

\subsubsection{Extending the pretrained model with Gaussian Mixture Variational Encoding}
\label{architecture}

Previous work has shown that explicitly modeling the latent embedding space of pretrained single-cell models as a clustered structure can improve cell representations~\cite{wen2023cellplm}. Motivated by this, we further augment the pretrained model with a Gaussian Mixture Variational Encoder (GMVE)~\cite{dilokthanakul2016deep}. Given the pretrained model output $\mathbf{h}_c = f_{\theta^{pt}}(\mathbf{c}_o) \in \mathbb{R}^{d}$, the GMVE defines the variational posterior $q(\mathbf{z}_c \mid \mathbf{h}_c)$,
\vspace{-0.1cm}
\begin{equation}
q(\mathbf{z}_c \mid \mathbf{h}_c) = \sum_{l=1}^{L} \pi_l(\mathbf{h}_c) \mathcal{N}(\mathbf{z}_c \mid \boldsymbol{\mu}_l(\mathbf{h}_c), \boldsymbol{\Sigma}_l(\mathbf{h}_c))
\end{equation}

where $\pi_l(\mathbf{h}_c)$ is the mixture weight ($\sum \pi_l = 1$), and $\boldsymbol{\mu}_l, \boldsymbol{\Sigma}_l$ are the parameters of the $l$-th Gaussian. To regularize this structure, we minimize the KL divergence between the variational posterior and a learned Gaussian mixture prior,
\vspace{-0.1cm}
\begin{equation}
\mathcal{L}_{\text{GMVE}} = \text{KL} \left( q(\mathbf{z}_c \mid \mathbf{h}_c) \parallel \sum_{l=1}^{L} p_l \mathcal{N}(\mathbf{z}_c \mid \boldsymbol{\mu}_l, \boldsymbol{\Sigma}_l) \right)
\end{equation}

\subsubsection{Overall post-pretraining objective}
\label{overall-loss}

Finally, the overall objective of \text{{\OurModel}} combines the regular pretraining loss $\mathcal{L}_{\text{pre}}$ (refer to Appendix~\ref{scrl-token-prediction} for preliminaries on pretraining loss), the prototype-guided regularization, the lineage-aware regularization, and the GMVE regularization.

\vspace{-0.1cm}
\begin{equation}
\mathcal{L}_{\text{Total}} = \mathcal{L}_{pre} + \lambda_1 \mathcal{L}_{prototype} + \lambda_2 \mathcal{L}_{\mathrm{lineage}} + \lambda_3 \mathcal{L}_{\mathrm{GMVE}}
\end{equation}
where \(\lambda_1\), \(\lambda_2\) and \(\lambda_3\) control the contributions of the three regularization terms.

\section{Experimental Setup}
\label{experiments}

\subsection{Evaluation Tasks}

We evaluate the utility of {\OurModel} on three widely used computational biology tasks: 1) cell identity prediction, 2) spatial transcriptomics imputation, and 3) gene perturbation response prediction. 

\textbf{Cell Identity Prediction}: Cell identity prediction is the task of assigning each cell a cell-type label from transcriptomic data. We define two versions of this task, on-domain and out-of-domain prediction. In the on-domain setting, both training and evaluation are performed on single-cell RNA sequencing (scRNA-seq) data. In the out-of-domain setting, models trained on scRNA-seq data are evaluated on spatial transcriptomics data to probe zero-shot out-of-distribution generalization.

We benchmark ten human single-cell transcriptomic datasets, including eight scRNA-seq datasets; peripheral blood \cite{zheng2017massively, gayoso2022python}, pancreas \cite{chen2023transformer}, liver \cite{lin2020scclassify}, myeloid \cite{cheng2021pan}, multiple sclerosis \cite{schirmer2019neuronal}, heart \cite{tucker2020transcriptional}, and lung \cite{kim2020single}, and two spatial transcriptomics datasets of liver \cite{10xgenomics2025datasets}. Detailed dataset descriptions can be found in Appendix~\ref{sec:dataset-details}. We report macro F1 scores for on-domain evaluations and recall@$k$ scores for zero-shot, out-of-domain transfer tasks; recall@$k$ denotes the proportion of test samples whose correct label appears among the top-$k$ retrieved candidates.

\textbf{Spatial Transcriptomic Imputation}: Spatial transcriptomic imputation is the task of predicting unmeasured gene expression from the subset of transcripts observed in each cell. We benchmark this task on the two spatial transcriptomics datasets. We randomly mask 15\% of genes as prediction targets and use the remaining genes as input. We evaluate performance using Pearson correlation and cosine similarity between predicted and ground-truth.

\textbf{Gene Perturbation Response Prediction}: Gene perturbation response prediction aims to predict changes in a cell's gene expression resulting from a specific perturbation, given the unperturbed gene expression profile as input. This task is key to developing pharmacological interventions. We formulate this as an out-of-distribution task by holding out one cell type from each reference dataset and using the held-out cell type as the query dataset. We tested this task on a peripheral blood dataset \cite{kang2018multiplexed}. We use Pearson correlation coefficient to measure the performance of this task.

\subsection{Baselines}
\label{sec:baselines}

We evaluated {\OurModel} against nine baselines for cell identity prediction, seven for spatial transcriptomics imputation, and three for zero-shot gene perturbation response prediction.

\section{Results}

In this section, we present the results of applying {\OurModel} to the three aforementioned tasks. We also provide an ablation study examining the contribution of each component of {\OurModel}, along with a qualitative analysis of the factors that may underlie its success in alignment.

\subsection{Cell Identity Prediction}

\begin{wrapfigure}{r}{0.45\columnwidth}
\vspace{-12pt}
  \centering
  \includegraphics[width=0.45\textwidth]{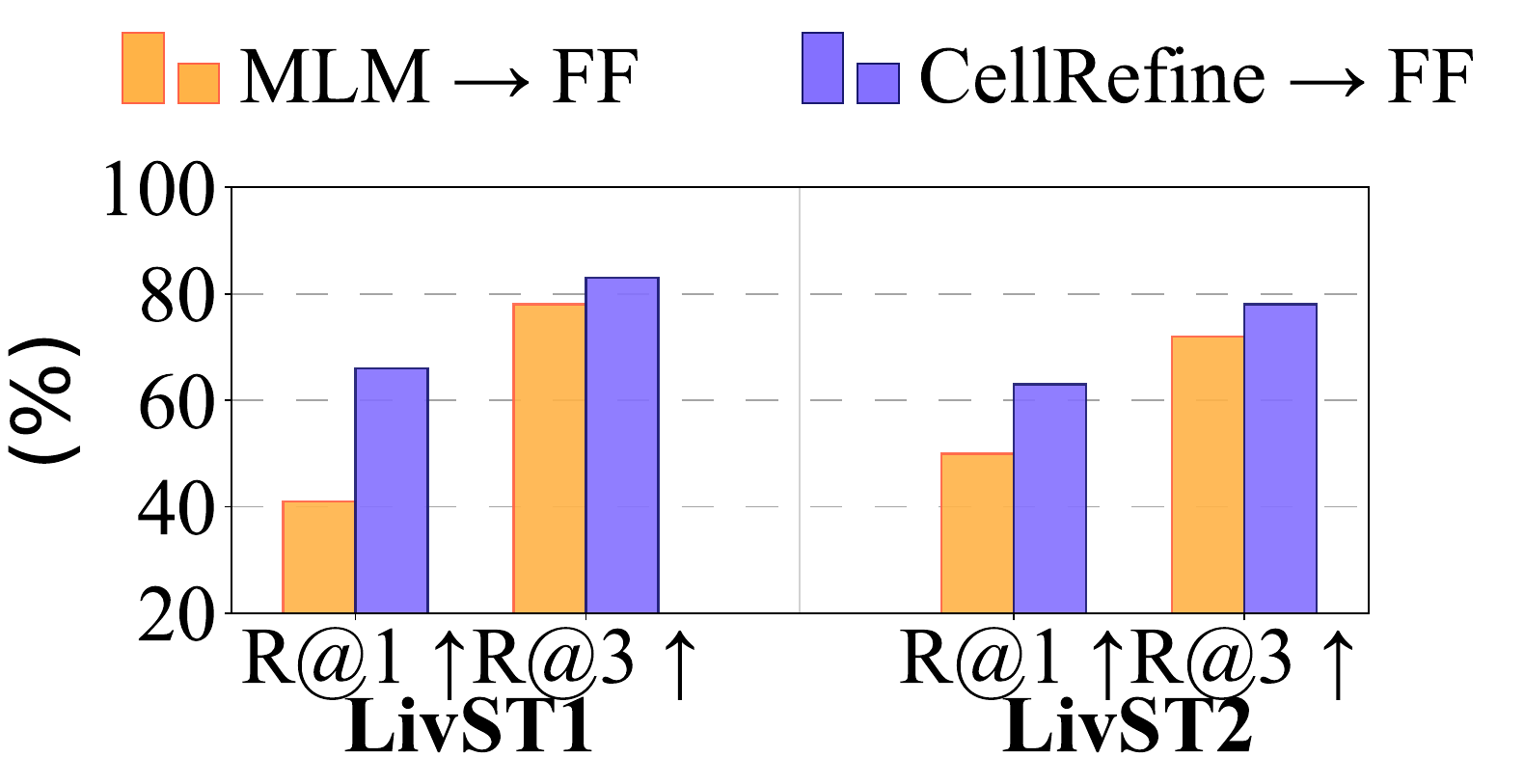}
  \caption{Out-of-domain zero-shot cell identity prediction performance (recall@k) on LivST1 and LivST2.}
  \label{fig:zero_shot_off_domain}
  \vspace{-12pt}
\end{wrapfigure}

\begin{table}[t]
\centering

\small
\setlength{\tabcolsep}{4pt}
\renewcommand{\arraystretch}{1.05}

\resizebox{\linewidth}{!}{
\begin{tabular}{lccccc}
\toprule
\textbf{Method}
& \textbf{Blood}
& \textbf{Myeloid}
& \textbf{Liver}
& \textbf{MS}
& \textbf{Avg.} $\uparrow$ \\
\midrule

PCA + LR
& $0.586 \pm 0.003$
& $0.340 \pm 0.001$
& $0.301 \pm 0.004$
& $0.708 \pm 0.004$
& $0.484 \pm 0.003$ \\

LR on HVG
& $0.723 \pm 0.007$
& $0.354 \pm 0.001$
& $0.323 \pm 0.002$
& $0.704 \pm 0.006$
& $0.526 \pm 0.004$ \\

scTOP
& $0.514 \pm 0.004$
& $0.251 \pm 0.000$
& $0.306 \pm 0.001$
& $0.624 \pm 0.001$
& $0.424 \pm 0.002$ \\

scTOP-rep on HVG + LR
& $0.739 \pm 0.006$
& $0.342 \pm 0.001$
& $0.340 \pm 0.001$
& $0.705 \pm 0.003$
& $0.532 \pm 0.003$ \\

\midrule

scANVI
& $0.632 \pm 0.004$
& $0.338 \pm 0.006$
& $0.321 \pm 0.008$
& $0.742 \pm 0.002$
& $0.508 \pm 0.005$ \\

CellTypist
& $0.686 \pm 0.005$
& $0.341 \pm 0.002$
& $0.281 \pm 0.004$
& $0.660 \pm 0.002$
& $0.492 \pm 0.003$ \\

\midrule

scBERT
& $0.704 \pm 0.011$
& $0.315 \pm 0.010$
& $0.207 \pm 0.021$
& $0.612 \pm 0.020$
& $0.460 \pm 0.016$ \\

\rowcolor{lightpurple}
scBERT + {\OurModel} (Ours)
& $0.817 \pm 0.012$
& $0.382 \pm 0.008$
& $0.365 \pm 0.010$
& $0.691 \pm 0.018$
& $0.564 \pm 0.012$ \\

\addlinespace[3pt]

scGPT
& $0.753 \pm 0.009$
& $0.342 \pm 0.007$
& $0.280 \pm 0.007$
& $0.701 \pm 0.012$
& $0.519 \pm 0.009$ \\

\rowcolor{lightpurple}
scGPT + {\OurModel} (Ours)
& $0.840 \pm 0.008$
& $0.377 \pm 0.011$
& $0.369 \pm 0.006$
& $0.740 \pm 0.008$
& $0.582 \pm 0.008$ \\

\addlinespace[3pt]

Geneformer
& $0.731 \pm 0.008$
& $0.354 \pm 0.006$
& $0.284 \pm 0.011$
& $0.719 \pm 0.009$
& $0.522 \pm 0.009$ \\

\rowcolor{lightpurple}
Geneformer + {\OurModel} (Ours)
& $\mathbf{0.843 \pm 0.005}$
& $\mathbf{0.391 \pm 0.006}$
& $\mathbf{0.382 \pm 0.008}$
& $\mathbf{0.754 \pm 0.003}$
& $\mathbf{0.593 \pm 0.006}$ \\

\bottomrule
\end{tabular}
}
\caption{On-domain cell identity prediction performance measured by macro-F1 (higher is better). Results are reported as mean $\pm$ standard deviation. Results of foundation models after applying {\OurModel} post-pretraining are highlighted in purple.}
\label{Cell identity prediction main results}
\vspace{-0.5cm}
\end{table}

In Table~\ref{Cell identity prediction main results}, we report on-domain cell identity prediction results, comparing \OurModel{} against a range of baselines across four datasets. Results for the remaining datasets, and cell class-level analysis are provided in Appendix~\ref{sec:additional-experiments-cip-benchmarks} and Appendix~\ref{sec:additional-experiments-class-level} respectively. Several findings stand out. First, post-pretraining with \OurModel{} consistently improves macro-F1 across all evaluated foundation-model backbones and datasets. In particular, \OurModel{} yields relative gains of more than 15\% over the corresponding Geneformer baseline on the Blood and Liver datasets, both of which exhibit pronounced class imbalance and substantial cross-dataset generalization challenges. Overall, the strongest non-foundation-model baseline, scTOP representations on HVGs with logistic regression, achieves an average macro-F1 of 0.532. Although all three foundation-model baselines initially fall below this level, post-pretraining with \OurModel{} raises the performance of scBERT, scGPT, and Geneformer beyond it. Among these variants, Geneformer + \OurModel{} achieves the best overall performance, with an average macro-F1 of 0.593.

Next, we examine out-of-domain cell identity prediction. Here, models are trained on scRNA-seq data as the source domain and evaluated zero-shot on spatial transcriptomics data as the target domain. Although spatial transcriptomics provides gene expression measurements together with the spatial coordinates of cells within tissue, labeled datasets in this modality remain limited. As a result, models must generalize across distribution shifts between sequencing-based and spatial measurement technologies. We therefore assess whether \text{\OurModel}, trained on scRNA-seq data, can improve cell identity prediction in spatial transcriptomics data without target-domain supervision.

Since Geneformer is the strongest-performing foundation-model baseline, we compare it with Geneformer + {\OurModel}. Recall@\(k\) for \(k \in \{1,3\}\) is reported in Figure~\ref{fig:zero_shot_off_domain}, where MLM \(\rightarrow\) FF denotes Geneformer and {\OurModel} \(\rightarrow\) FF denotes Geneformer + {\OurModel}. Geneformer + {\OurModel} consistently demonstrates stronger zero-shot generalization across both datasets, with notable gains in Recall@1. These results suggest that {\OurModel} improves the latent cell embedding space, yielding more robust representations that may reduce noise and enhance transfer across modalities.

\subsection{Spatial Transcriptomics Imputation}

\begin{table}[t]
\centering
\small
\setlength{\tabcolsep}{4pt}
\renewcommand{\arraystretch}{1.05}

\resizebox{\linewidth}{!}{
\begin{tabular}{lcccccc}
\toprule
\textbf{Method}
& \multicolumn{2}{c}{\textbf{LiverST1}}
& \multicolumn{2}{c}{\textbf{LiverST2}}
& \multicolumn{2}{c}{\textbf{Avg.}} \\
\cmidrule(lr){2-3}
\cmidrule(lr){4-5}
\cmidrule(lr){6-7}
& \textbf{Cos} $\uparrow$
& \textbf{Corr} $\uparrow$
& \textbf{Cos} $\uparrow$
& \textbf{Corr} $\uparrow$
& \textbf{Cos} $\uparrow$
& \textbf{Corr} $\uparrow$ \\
\midrule

SpaGE
& $0.331 \pm 0.012$
& $0.192 \pm 0.018$
& $0.368 \pm 0.009$
& $0.160 \pm 0.010$
& $0.350 \pm 0.011$
& $0.176 \pm 0.014$ \\

stPlus
& $0.382 \pm 0.014$
& $0.172 \pm 0.011$
& $0.291 \pm 0.009$
& $0.172 \pm 0.016$
& $0.337 \pm 0.012$
& $0.172 \pm 0.014$ \\

Tangram
& $0.263 \pm 0.005$
& $0.114 \pm 0.007$
& $0.301 \pm 0.008$
& $0.166 \pm 0.010$
& $0.282 \pm 0.007$
& $0.140 \pm 0.009$ \\

gimVI
& $0.312 \pm 0.010$
& $0.143 \pm 0.011$
& $0.327 \pm 0.010$
& $0.171 \pm 0.009$
& $0.320 \pm 0.010$
& $0.157 \pm 0.010$ \\

\midrule

scBERT
& $0.392 \pm 0.012$
& $0.288 \pm 0.013$
& $0.421 \pm 0.019$
& $0.261 \pm 0.009$
& $0.407 \pm 0.016$
& $0.275 \pm 0.011$ \\

\rowcolor{lightpurple}
scBERT + {\OurModel} (Ours)
& $0.502 \pm 0.016$
& $0.300 \pm 0.012$
& $0.487 \pm 0.017$
& $0.289 \pm 0.012$
& $0.495 \pm 0.017$
& $0.295 \pm 0.012$ \\

\addlinespace[3pt]

scGPT
& $0.421 \pm 0.011$
& $0.291 \pm 0.009$
& $0.442 \pm 0.013$
& $0.310 \pm 0.016$
& $0.432 \pm 0.012$
& $0.301 \pm 0.013$ \\

\rowcolor{lightpurple}
scGPT + {\OurModel} (Ours)
& $0.494 \pm 0.013$
& $0.311 \pm 0.009$
& $0.502 \pm 0.011$
& $0.330 \pm 0.011$
& $0.498 \pm 0.012$
& $0.321 \pm 0.010$ \\

\addlinespace[3pt]

Geneformer
& $0.477 \pm 0.011$
& $0.303 \pm 0.010$
& $0.491 \pm 0.012$
& $0.322 \pm 0.009$
& $0.484 \pm 0.012$
& $0.313 \pm 0.010$ \\

\rowcolor{lightpurple}
Geneformer + {\OurModel} (Ours)
& $\mathbf{0.526 \pm 0.013}$
& $\mathbf{0.313 \pm 0.011}$
& $\mathbf{0.514 \pm 0.008}$
& $\mathbf{0.342 \pm 0.010}$
& $\mathbf{0.520 \pm 0.011}$
& $\mathbf{0.328 \pm 0.011}$ \\

\bottomrule
\end{tabular}
}
\caption{Spatial transcriptomics imputation performance measured by cosine similarity (Cos) and Pearson correlation (Corr), higher is better. Results are reported as mean $\pm$ standard deviation.}
\label{st_imputation}
\end{table}

For the spatial transcriptomics imputation task, we evaluate four non-foundation-model baselines and three foundation models, scBERT, scGPT, and Geneformer, along with their corresponding variants post-pretrained with {\OurModel}. The foundation-model baselines generally outperform the non-foundation-model methods, with Geneformer achieving the strongest baseline performance. Consistent with the trend observed in cell identity prediction, post-pretraining with {\OurModel} further improves all three foundation-model backbones. The results are reported in Table~\ref{st_imputation}.

\subsection{Gene Perturbation Response Prediction}

\begin{table}[t]
\centering

\small
\setlength{\tabcolsep}{5pt}
\renewcommand{\arraystretch}{1.05}

\resizebox{\linewidth}{!}{
\begin{tabular}{lccccc}
\toprule
\textbf{Method}
& \textbf{CD4 T}
& \textbf{CD14+ Mono}
& \textbf{CD8 T}
& \textbf{Dendritic}
& \textbf{Avg.} $\uparrow$ \\
\midrule

scBERT
& $0.198 \pm 0.003$
& $0.518 \pm 0.005$
& $0.228 \pm 0.007$
& $0.421 \pm 0.004$
& $0.341 \pm 0.005$ \\

\rowcolor{lightpurple}
scBERT + {\OurModel} (Ours)
& $0.214 \pm 0.003$
& $0.530 \pm 0.002$
& $0.232 \pm 0.006$
& $0.467 \pm 0.002$
& $0.361 \pm 0.003$ \\

\addlinespace[3pt]

scGPT
& $0.210 \pm 0.006$
& $0.525 \pm 0.004$
& $0.229 \pm 0.002$
& $0.450 \pm 0.006$
& $0.354 \pm 0.005$ \\

\rowcolor{lightpurple}
scGPT + {\OurModel} (Ours)
& $\mathbf{0.234 \pm 0.008}$
& $0.539 \pm 0.003$
& $0.243 \pm 0.002$
& $0.481 \pm 0.007$
& $0.374 \pm 0.005$ \\

\addlinespace[3pt]

Geneformer
& $0.204 \pm 0.002$
& $0.527 \pm 0.003$
& $0.231 \pm 0.001$
& $0.452 \pm 0.001$
& $0.354 \pm 0.002$ \\

\rowcolor{lightpurple}
Geneformer + {\OurModel} (Ours)
& $0.232 \pm 0.002$
& $\mathbf{0.542 \pm 0.001}$
& $\mathbf{0.245 \pm 0.001}$
& $\mathbf{0.485 \pm 0.001}$
& $\mathbf{0.376 \pm 0.001}$ \\

\bottomrule
\end{tabular}
}
\caption{Zero-shot gene perturbation response prediction results. Variants of {\OurModel} are highlighted in purple, and the best overall result for each dataset is shown in bold.}
\label{perturbation_response_prediction}
\vspace{-0.5cm}
\end{table}

The results for gene perturbation response prediction are shown in Table~\ref{perturbation_response_prediction}. This task is widely regarded as challenging, and prior studies have reported relatively low performance overall~\cite{kendiukhov2026sparse}. Despite this difficulty, \text{\OurModel} yields consistent improvements over all three initial foundation model baselines, with scGPT + {\OurModel} leading in the cell type CD4 T and Geneformer + {\OurModel} achieving the overall strongest performance. 

Additional result comparisons with parameter-efficient methods are provided in Appendix~\ref{sec:additional-experiments-baselines}. 

\subsection{Why \text{\OurModel} Works?}

\begin{figure}[t]
    \centering
    \label{umaps}
    \begin{subfigure}[t]{0.48\textwidth}
        \includegraphics[width=\textwidth]{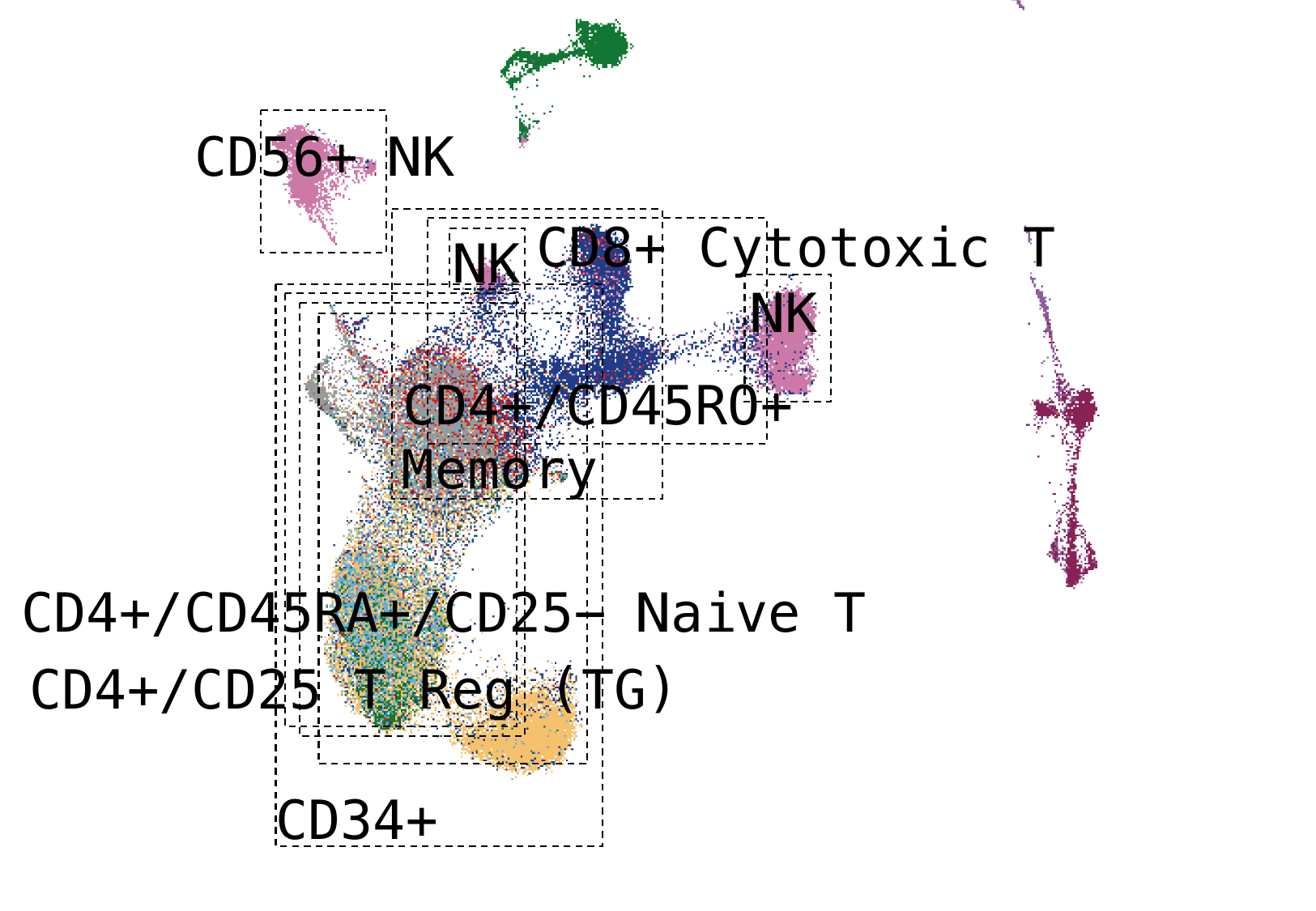}
        \caption{UMAP visualization of blood cell embeddings after applying MLM without \text{\OurModel} for alignment. Silhouette: 0.445, ARI: 0.258}
        \label{fig:umap1a}

    \end{subfigure}
    \hfill
    \begin{subfigure}[t]{0.48\textwidth}
        \includegraphics[width=\textwidth]{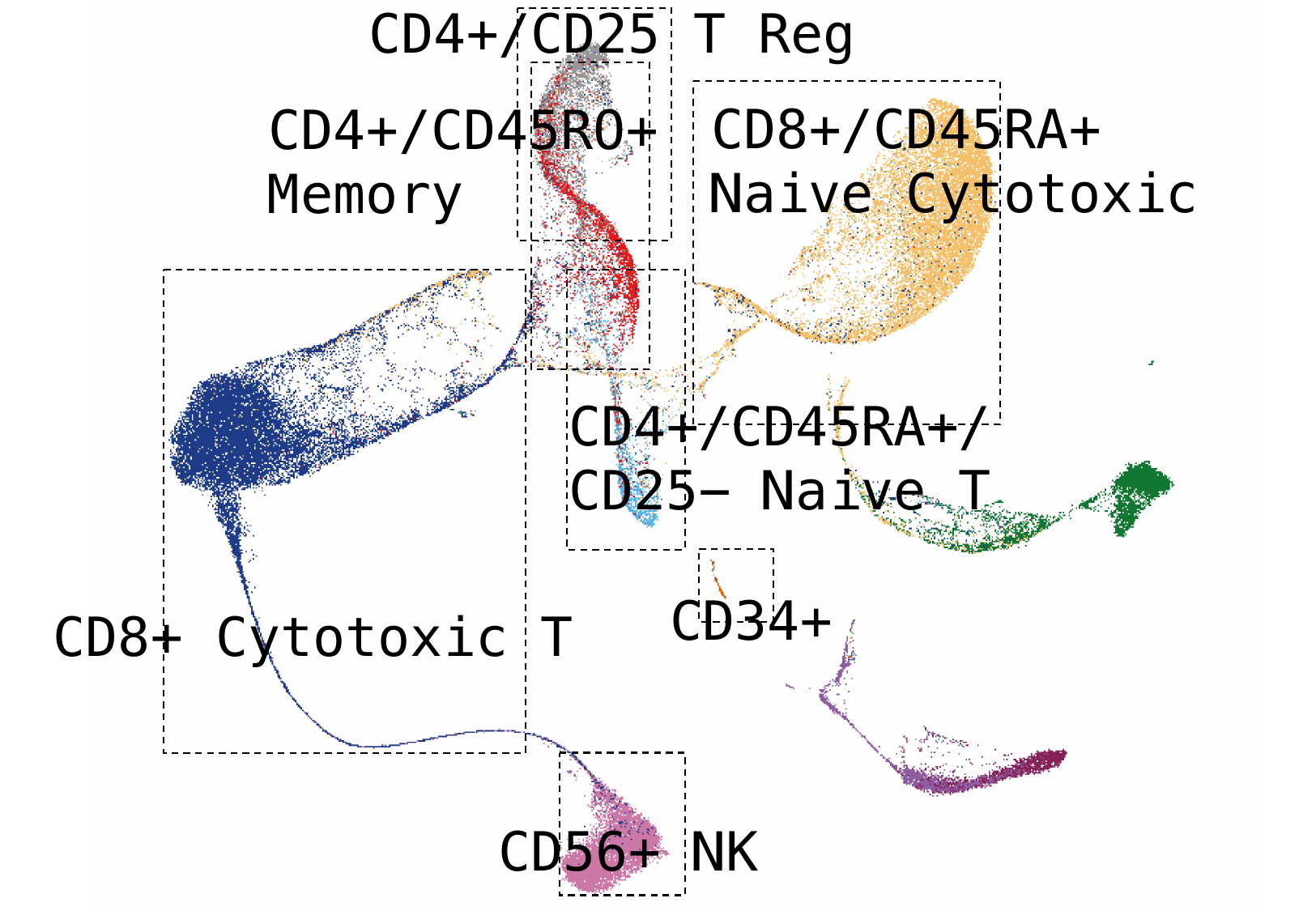}
        \caption{UMAP visualization of blood cell embeddings after post-pretraining with \text{\OurModel} for alignment. Silhouette: 0.695, ARI: 0.611}
        \label{fig:umap1b}

    \end{subfigure}
    \caption{UMAP visualizations of latent Geneformer cell embeddings on the Blood dataset~\cite{zheng2017massively} before (Figure~\ref{fig:umap1a}) and after (Figure~\ref{fig:umap1b}) post-pretraining with \OurModel{}. Cell-type clusters are outlined with dashed lines. After applying \OurModel{}, the embeddings exhibit improved separation between cell clusters compared with the original Geneformer representations. The number of neighbors and the minimum distance used in generating UMAPs are 15 and 0.1, respectively.}
    \vspace{-0.5115cm}
\end{figure}

In this section, we qualitatively analyze the latent embedding space of Geneformer, which shows the strongest performance, before and after applying \text{\OurModel}, with the goal of gaining insight into the changes induced by \text{\OurModel} and why these changes may lead to improved overall performance, specifically with respect to the rare cells in the tail of the cell distribution. To this end, we use the Blood dataset~\cite{zheng2017massively} and extract latent cell embeddings both before and after applying \text{\OurModel}. We then use UMAP to project these embeddings into a two-dimensional space for visualization. The resulting plots are shown in Figure~\ref{fig:umap1a} (before applying \text{\OurModel}) and Figure~\ref{fig:umap1b} (after applying \text{\OurModel}), with cell clusters annotated with dashed lines. 
\vspace{-0.1cm}

Figures~\ref{fig:umap1a} and~\ref{fig:umap1b} suggest that \text{\OurModel} substantially reorganizes the latent embedding space, making underrepresented cell types more distinguishable. Prior to applying \text{\OurModel}, as shown in Figure~\ref{fig:umap1a}, tail cells exhibit substantial overlap with major cell populations. After refinement, Figure~\ref{fig:umap1b} shows distinct cell regions with clearer boundaries. This indicates that \text{\OurModel} promotes a more structured representation space in which underrepresented cell types are less obscured by dominant populations, better supporting downstream fine-tuning and explaining improvements across the three benchmark tasks. Additional visualizations of other datasets are given in Appendix~\ref{subsec:latent-space}.

\vspace{-0.1cm}
\subsection{Effectiveness of the Multi-Term Objective}

\begin{wraptable}{r}{0.45\linewidth}
\centering
\small
\setlength{\tabcolsep}{3pt}
\resizebox{\linewidth}{!}{%
\begin{tabular}{lccc}
\toprule
\textbf{Loss Terms Used} & \textbf{CIP } & \textbf{STI} & \textbf{PRP} \\
\midrule
$\mathcal{L}_{\text{pre}}$ 
& 0.522 & 0.484 & 0.354 \\
$\mathcal{L}_{\text{pre}} + \mathcal{L}_{C \rightleftarrows P}$ 
& 0.568 & 0.501 & 0.362 \\
$\mathcal{L}_{\text{pre}} + \mathcal{L}_{C \rightleftarrows P} + \mathcal{L}_{\mathrm{lineage}}$ 
& 0.572 & 0.511 & 0.368 \\
$\mathcal{L}_{\text{pre}} + \mathcal{L}_{C \rightleftarrows P} + \mathcal{L}_{\mathrm{lineage}} + \mathcal{L}_{\mathrm{GMVE}}$ 
& 0.593 & 0.520 & 0.376 \\
\bottomrule
\end{tabular}
}
\caption{Ablation study on the effectiveness of each loss term in \text{\OurModel} for cell identity prediction (CIP) (macro F1), spatial transcriptomics imputation (STI) (cosine similarity), and perturbation response prediction (PRP) (Pearson correlation coefficient).}
\label{tab:ablation-loss}
\end{wraptable}

As described in Section~\ref{overall-loss}, \text{\OurModel} optimizes four loss terms that jointly shape the learned representation space. In this section, we perform an ablation study to evaluate the contribution of each term to the overall performance by progressively adding the loss terms one at a time. In Table~\ref{tab:ablation-loss}, we report average cell classification performance of the datasets in Table ~\ref{Cell identity prediction main results}, average spatial transcriptomics imputation on liver spatial transcriptomic datasets in Table ~\ref{st_imputation}, and average perturbation response prediction across cell types in the perturbed peripheral blood dataset in Table \ref{perturbation_response_prediction}. In all ablation experiments, we use Geneformer as the backbone, which is the strongest-performing foundation model in our evaluations. The progressive gains observed as additional objectives are introduced suggest that the loss terms provide complementary learning signals during post-pretraining. As the table indicates, \text{\OurModel} achieves the best performance when all four loss terms are included, confirming that each component of the multi-objective contributes to improving the embedding space and that their combination is particularly effective.

Additional ablation experiments isolating the contribution of marker biology are provided in Appendix~\ref{sec:additional-experiments-contribution-marker-biology}. We further report a sensitivity analysis of the loss-function hyperparameters in Appendix~\ref{sec:loss-hp} and a comparison of computational overhead in Appendix~\ref{sec:compute-overhead}.

\section{Related Work}

In this section, we review related work, organized along a few key areas of interest.

\textbf{Adaptation between pretraining and fine-tuning}: Inserting a second pretraining stage before task-specific fine-tuning is established practice. Domain and task-adaptive pretraining continue the masked language modeling objective on in-domain data \cite{gururangan2020don}, and post-pre-training has recently been used to realign vision–language models prior to downstream adaptation \cite{yamaguchi2025post}. Although {\OurModel} also introduces an intermediate stage, the role of that stage is different. Continued MLM reuses the stage-one objective, which moves the embedding manifold toward the target distribution but imposes no structure the original objective did not already encode. {\OurModel} modifies the objective itself, adding terms defined over external biological structure rather than over observed tokens.

\textbf{Prototype and ontology-guided representation learning}: Anchoring representations to class-level embeddings is an established alternative to a free classification head, as in prototypical networks \cite{snell2017prototypical} and in contrastive image–text pretraining, where a text encoder supplies the class anchors \cite{radford2021learning}. In single-cell analysis this has been realized through textual descriptions of cell types, as in LangCell \cite{zhao2024langcell}, GenePT \cite{chen2025simple}, and Cell2Sentence \cite{levine2024cell2sentence}. {\OurModel} differs in both the anchor and the stage at which it is applied. The anchor is an ontology-ordered marker gene program $\mathcal{H}(c_i)$ rather than free text, expressed in a form the pretrained tokenizer accepts directly and carrying the discriminative signal practitioners already use for manual annotation \cite{dumitrascu2021optimal}. Both the prototype and lineage terms act as post-pretraining regularizers rather than classification objectives, so they constrain the representation without committing to a downstream head. In this sense, \OurModel is agnostic to the fine-tuning method that follows.

\textbf{Long-tailed and shift-robust learning}: Long-tailed recognition is typically addressed by reweighting or margin adjustment \cite{cui2019class, cao2019learning}, by logit adjustment under the class prior \cite{menon2020long}, or by decoupling representation learning from classifier learning \cite{kang2019decoupling}.  Within single-cell analysis, adaptive resampling has been proposed for imbalanced datasets \cite{navidi2025adaptive}. Covariate shift is usually treated separately, through batch integration methods such as Harmony \cite{korsunsky2019fast} and scANVI \cite{xu2021probabilistic}. {\OurModel} differs from both families in what it modifies. Reweighting and resampling methods require labeled tail examples and act primarily on the classifier, leaving the representation that classifier reads largely unchanged, while batch integration methods operate on expression values without access to cell-type semantics. {\OurModel} instead reshapes the representation using structure that exists for rare cell types even when samples do not, since marker programs and ontology relations are curated independently of any dataset. The same mechanism addresses both failure modes, because a marker-derived prototype is a target that neither depends on cell-type frequency nor varies with technical effects that leave marker identity intact.

\section{Conclusion and Future Work}
\label{conclusion}

In this work, we introduce \text{\OurModel}, a post-pretraining approach for improving single-cell pretrained models prior to fine-tuning. Unlike conventional pipelines that rely solely on pretraining followed by task-specific adaptation, \text{\OurModel} incorporates biological priors during an intermediate optimization stage, refining cell embeddings and improving robustness to long-tail induced data imbalance and covariate shift. Across diverse computational biology tasks, \text{\OurModel} consistently improves downstream performance.

These results position \text{\OurModel} as a promising direction for single-cell representation learning and highlight the value of integrating biological domain knowledge into large-scale representation learning frameworks. A limitation of the current approach is its reliance on established marker knowledge, which makes it less suitable for novel or poorly characterized cell types. Future work can therefore explore ways to reduce this dependence and extend biologically guided post-pretraining to settings involving previously uncharacterized populations. The intended regime of \text{\OurModel} is instead cell types that are well characterized at the marker level but remain difficult to model because sufficiently large, high-quality labeled datasets are hard to obtain. Such difficulty can arise from rare biology, labeling ambiguity, and technical artifacts, all of which are largely orthogonal to whether marker knowledge is available. This is precisely the long-tailed and closely related cell-type regime in which the gains from \text{\OurModel} are most pronounced.

\subsection*{AI use statement}

In this work, we used generative AI tools for parts of the method implementation, primarily to edit software code.
We have not used generative AI tools to help develop theoretical models or conceptual frameworks, formulate mathematical claims, propose or refine hypotheses, design or provide feedback on research methodology or experiments, implement methods (full implementation), assist with translation, clean and reformat dataset, support qualitative and thematic data analysis, interpret results, and generating synthetic data sets, providing critical ingredients for proving mathematical claims, assisting in the writing of proofs are not applicable to this work.
Additionally, we used generative AI tools to edit the research paper to improve readability, format tables and figures, edit software code, identify relevant literature, and searching for information. We have reviewed all AI-assisted work. All LLM-generated software code was verified and tested for correctness by 2 authors. We conducted manual literature survey of LLM-suggested related work in the domain, and checked all figures and tables formatted by LLMs to make sure the contents were not distorted during formatting. We take responsibility for the final content of this work,
including text, claims or artifacts produced with the aid of generative AI.

\subsection*{Ethics statement}

This work does not involve human-subject recruitment, personally identifiable information, or the release of new datasets. All experiments are conducted using existing publicly available benchmark datasets and computational methods. We do not identify any additional ethical concerns specific to this work.

\subsection*{Reproducibility statement}

We provide detailed descriptions of the model architecture, training objectives, experimental setup, datasets, evaluation protocols, and hyperparameters in the main paper and appendix to facilitate reproducibility. Additional implementation details and ablation settings are also included in the appendix. We will publicly release the source code and necessary scripts upon acceptance.

\bibliography{iclr2027_conference}
\bibliographystyle{iclr2027_conference}

\clearpage
\appendix


\phantomsection
\label{app:contents}
\begin{center}
{\Huge\bfseries Appendix\par}
\end{center}
\vspace{0.5em}
\startcontents[appendix]
\printcontents[appendix]{}{1}{\setcounter{tocdepth}{2}}
\clearpage

\section{Broader Impacts}
\label{sec:broader-impacts}

Advances in single-cell representation learning, such as the framework proposed in this work, have the potential to significantly accelerate progress in biomedical research and healthcare. By enabling more accurate and robust characterization of cellular identities, particularly for rare and clinically relevant cell populations, our approach may contribute to improved understanding of disease mechanisms, earlier detection of pathological states, and the development of more targeted therapeutic interventions. In particular, better modeling of rare cell types could facilitate breakthroughs in areas such as cancer biology, immunology, and regenerative medicine, where small but critical cell populations often drive disease progression or treatment response.

Beyond direct clinical applications, our method may also enhance the utility of large-scale single-cell atlases and public datasets by improving cross-dataset generalization in the presence of technical variability. This could promote more reproducible and integrative analyses across research groups, enabling more efficient knowledge sharing and collaborative discovery. Furthermore, incorporating biologically grounded inductive biases into machine learning models may encourage a broader shift toward hybrid approaches that combine data-driven learning with domain expertise, potentially improving interpretability and trustworthiness in computational biology.

However, several potential risks and limitations should be considered. First, while our approach aims to better capture rare cell populations, biases in the underlying datasets—such as underrepresentation of certain populations, disease states, or demographic groups—may still propagate into the learned representations. This could lead to uneven performance across biological contexts and limit the generalizability of the model in real-world clinical settings. Second, reliance on curated biological knowledge, such as marker gene databases and cell ontologies, introduces dependencies on the completeness and accuracy of these resources. Errors or omissions in these databases could bias the model or reinforce existing misconceptions in the literature.

Additionally, as with many machine learning models applied to biomedical data, there are considerations related to data privacy and ethical use. Although our work focuses on aggregated gene expression data rather than identifiable patient information, downstream applications in clinical settings may involve sensitive datasets, requiring careful handling, appropriate consent, and adherence to regulatory standards.

Finally, while improved predictive performance can support scientific discovery, such models should not be used in isolation for clinical decision-making without rigorous validation. Ensuring transparency, interpretability, and proper benchmarking across diverse datasets will be essential for responsible deployment.

Overall, this work contributes toward more biologically informed and robust machine learning methods for single-cell analysis, with the potential for meaningful positive impact in biomedical research and healthcare, while highlighting the importance of careful consideration of data biases, knowledge dependencies, and ethical use. 

\section{Gene marker set and prototype construction}
\label{marker construction}

For each cell in the post-pretraining cohort, we construct a comprehensive set of marker genes by aggregating curated human cell type markers from multiple high-quality biological databases. Specifically, we integrate marker gene information from the MSigDB C8 collection \cite{subramanian2005gene, liberzon2011molecular}, CellMarker 2.0 \cite{hu2023cellmarker}, and PanglaoDB \cite{franzen2019panglaodb}. These resources provide complementary coverage of cell type–specific gene signatures across a wide range of tissues and biological contexts. By combining these sources, we obtain a robust and diverse marker gene set for each cell type, reducing reliance on any single database and improving coverage of both well-characterized and less-studied cell populations.

Once the marker gene sets are compiled, we organize them according to their positions within the cell ontology using a marker organization function \(H(c_i)\), where \(c_i\) denotes a specific cell type. The purpose of this function is to impose a biologically meaningful ordering on the marker genes that reflects the hierarchical structure of the cell ontology. Rather than treating marker genes as an unordered set, \(H(c_i)\) leverages the ontology to encode relationships between cell types at different levels of granularity.

In particular, \(H(c_i)\) orders marker genes based on their specificity within the ontology hierarchy. Genes associated with highly specific cell identities (e.g., fine-grained sub-subtypes) are given higher priority and placed earlier in the sequence. These are followed by genes characteristic of intermediate cell subtypes, and finally by genes associated with broader, high-level cell categories. This ordering captures a progression from fine-grained to coarse-grained biological identity, enabling the representation to emphasize discriminative features of closely related cell types while still retaining information about broader lineage relationships.

For example, consider a marker gene set \(\{A, B, C\}\) for a given cell type, where gene \(A\) is associated with a major cell type, gene \(B\) corresponds to an intermediate cell subtype, and gene \(C\) is specific to a more refined cell sub-subtype. Applying the organization function \(H(c_i)\) reorders this set into the sequence \((C, B, A)\), reflecting a hierarchy from the most specific to the most general markers. 

We refer to this ordered sequence of marker genes as the \textit{prototype} of the cell type. This prototype serves as a structured representation that encapsulates both the defining molecular features of the cell and their hierarchical relationships within the broader cell ontology.

\begin{figure}[h]
  \centering
  \includegraphics[width=1\linewidth]{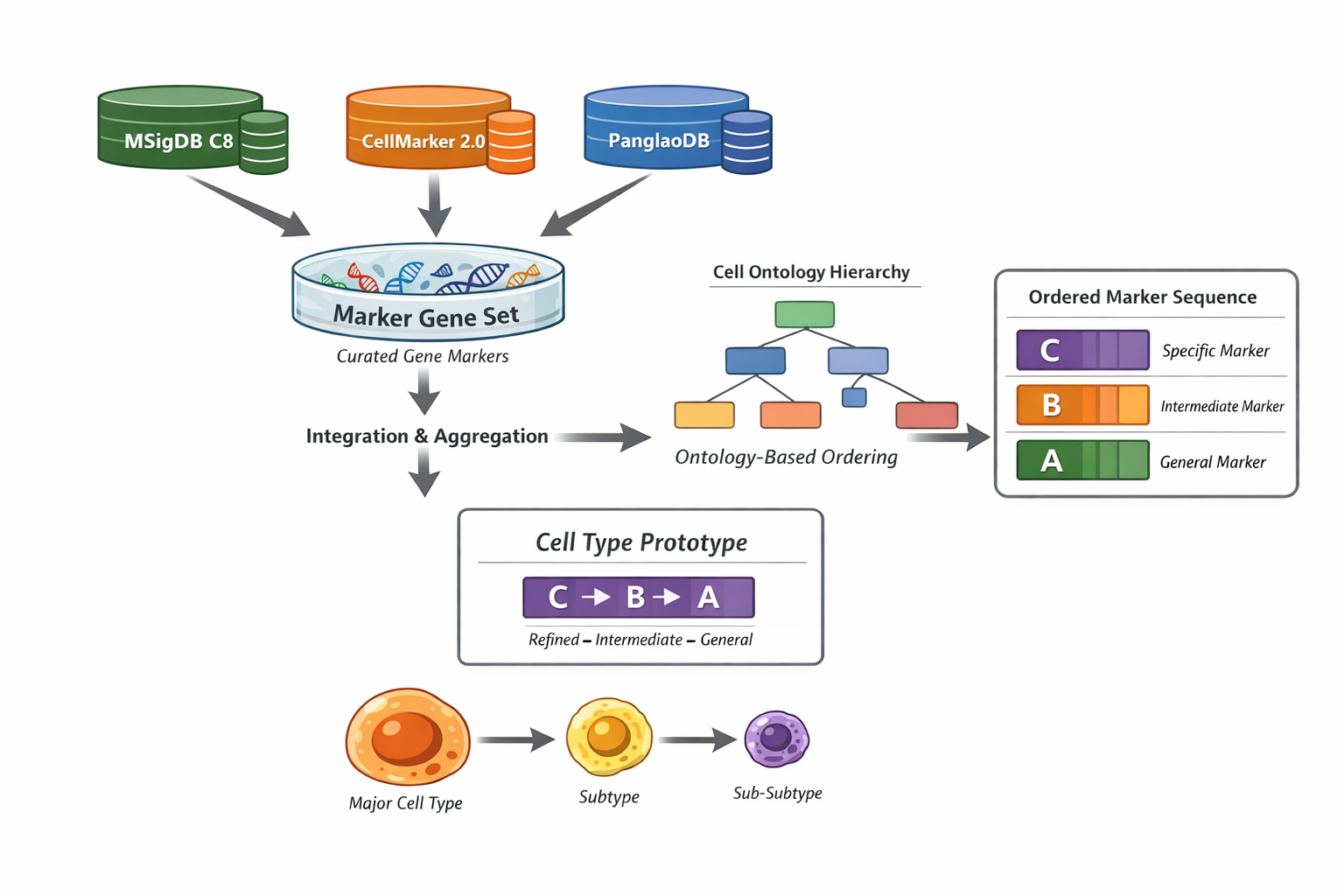}
    \caption{Marker gene set and cell type prototype creation procedure.}
  \label{gene_marker_creation}
\end{figure}

\section{SCRL as Language Prediction Tasks}
\label{scrl-token-prediction}

Many pretrained models proposed for SCRL employ self-supervised token prediction objectives, including masked language modeling and autoregressive-style prediction~\cite{theodoris2023transfer}. These objectives provide a natural mechanism for learning dependencies among genes and their expression patterns from large collections of unlabeled single-cell data. In this section, we formulate these two common pretraining paradigms and describe their application to SCRL.

We begin by representing each cell as a sequence of gene tokens drawn from a fixed vocabulary
$\mathcal{V} = \{x_1,\dots,x_k\}$. For each cell $c$, a deterministic tokenization procedure $\tau(\cdot)$ maps its gene expression profile into an ordered token sequence of length $t$,

\begin{equation}
\mathbf{c} = \tau(c) = (x_1,\dots,x_t),
\qquad x_i \in \mathcal{V}.
\end{equation}

\paragraph{Masked language modeling.}
In masked language modeling, a subset of token indices
$M \subseteq \{1,\dots,t\}$ is masked from the input. Let

\begin{equation}
\mathbf{c}_{\mathrm{o}}
=
(x_i : i \notin M)
\end{equation}

denote the observed tokens. The objective is to learn a parameterized model
$f_\theta$ that captures the conditional distribution of each masked token given the observed context,

\begin{equation}
f_\theta :
\mathbf{c}_{\mathrm{o}}
\longmapsto
p_\theta(x_i \mid \mathbf{c}_{\mathrm{o}}),
\qquad i \in M.
\end{equation}

The model parameters $\theta$ are optimized by minimizing the masked-token-modeling objective

\begin{equation}
\mathcal{L}_{\mathrm{MLM}}
=
-\sum_{i \in M}
\log p_\theta(x_i \mid \mathbf{c}_{\mathrm{o}})
+
\alpha \mathcal{L}(\theta),
\label{l_mlm}
\end{equation}

where $\mathcal{L}(\theta)$ denotes an optional regularization term and $\alpha$ controls its contribution to the objective. In our experiments, we use $\mathcal{L}_{\mathrm{MLM}}$ as $\mathcal{L}_{\mathrm{pre}}$ for scBERT and Geneformer, following their original pretraining schemes. 

\paragraph{Autoregressive token prediction.}
An alternative is to learn the cell representation through autoregressive prediction. Given an ordered token sequence
$\mathbf{c}=(x_1,\dots,x_t)$, the joint probability of the sequence can be factorized as

\begin{equation}
p_\theta(\mathbf{c})
=
\prod_{i=1}^{t}
p_\theta(x_i \mid x_{<i}),
\end{equation}

where
$x_{<i}=(x_1,\dots,x_{i-1})$
denotes the tokens preceding position $i$. The model is therefore trained to predict each token conditioned on the previously observed tokens,

\begin{equation}
f_\theta :
x_{<i}
\longmapsto
p_\theta(x_i \mid x_{<i}).
\end{equation}

The corresponding autoregressive objective is

\begin{equation}
\mathcal{L}_{\mathrm{AR}}
=
-\sum_{i=1}^{t}
\log p_\theta(x_i \mid x_{<i})
+
\alpha \mathcal{L}(\theta).
\label{l_ar}
\end{equation}

In our experiments, we use $\mathcal{L}_{\mathrm{AR}}$ as $\mathcal{L}_{\mathrm{pre}}$ for scGPT, following it's original pretraining schemes.

Variants of this formulation are used by generative single-cell foundation models, where specialized attention masks can define which genes or expression values are treated as observed and which are predicted. Thus, although the precise prediction target and masking strategy may differ across models, both masked and autoregressive objectives train the model by recovering cellular information from partial context.

In both cases, the self-supervised prediction objective encourages the model to capture statistical dependencies among genes and gene-expression patterns. The resulting internal representations can subsequently be adapted to downstream SCRL tasks such as cell identity prediction, spatial transcriptomics imputation, and perturbation-response prediction.

\section{Additional Experiments and Results}
\label{sec:additional-experiments}

\subsection{Cell Identity Prediction Performance on Additional Benchmark Datasets}
\label{sec:additional-experiments-cip-benchmarks}

\begin{table}[t]
\centering

\small
\setlength{\tabcolsep}{5pt}
\renewcommand{\arraystretch}{1.05}

\begin{tabular}{lcccc}
\toprule
\textbf{Method}
& \textbf{Pancreas}
& \textbf{PBMC10K}
& \textbf{Heart}
& \textbf{Overall} $\uparrow$ \\
\midrule

PCA + LR
& $0.809 \pm 0.009$
& $0.954 \pm 0.007$
& $0.871 \pm 0.005$
& $0.878 \pm 0.007$ \\

LR on HVG
& $0.822 \pm 0.003$
& $0.953 \pm 0.007$
& $0.803 \pm 0.005$
& $0.859 \pm 0.005$ \\

scTOP
& $0.842 \pm 0.000$
& $0.924 \pm 0.014$
& $0.846 \pm 0.004$
& $0.871 \pm 0.006$ \\

scTOP-rep on HVG + LR
& $0.826 \pm 0.006$
& $0.954 \pm 0.004$
& $0.887 \pm 0.004$
& $0.889 \pm 0.005$ \\

\midrule

scANVI
& $0.704 \pm 0.007$
& $0.902 \pm 0.003$
& $0.859 \pm 0.006$
& $0.822 \pm 0.005$ \\

CellTypist
& $0.713 \pm 0.001$
& $0.886 \pm 0.009$
& $0.864 \pm 0.004$
& $0.821 \pm 0.005$ \\

\midrule

scBERT
& $0.692 \pm 0.007$
& $0.915 \pm 0.004$
& $0.841 \pm 0.007$
& $0.816 \pm 0.006$ \\

\rowcolor{lightpurple}
scBERT + \OurModel
& $0.811 \pm 0.007$
& $0.946 \pm 0.007$
& $0.906 \pm 0.011$
& $0.888 \pm 0.008$ \\

\addlinespace[3pt]

scGPT
& $0.716 \pm 0.010$
& $0.942 \pm 0.003$
& $0.849 \pm 0.010$
& $0.836 \pm 0.008$ \\

\rowcolor{lightpurple}
scGPT + \OurModel
& $\mathbf{0.842 \pm 0.007}$
& $0.953 \pm 0.014$
& $0.907 \pm 0.007$
& $\mathbf{0.901 \pm 0.009}$ \\

\addlinespace[3pt]

Geneformer
& $0.721 \pm 0.007$
& $0.951 \pm 0.007$
& $0.865 \pm 0.006$
& $0.846 \pm 0.007$ \\

\rowcolor{lightpurple}
Geneformer + \OurModel
& $0.829 \pm 0.005$
& $\mathbf{0.963 \pm 0.007}$
& $\mathbf{0.912 \pm 0.005}$
& $\mathbf{0.901 \pm 0.006}$ \\

\bottomrule
\end{tabular}
\caption{Cell identity prediction performance measured by macro-F1 (higher is better). Results are reported as mean $\pm$ standard deviation.}
\label{tab:cell_identity_appendix}

\end{table}

In Table~\ref{tab:cell_identity_appendix}, we report cell identity prediction performance on three additional benchmark datasets: Pancreas, PBMC10K, and Heart. Among the baselines, logistic regression applied to scTOP representations on HVGs achieves the strongest overall performance, while all three foundation-model baselines initially fall below it. Post-pretraining with \OurModel{}, however, consistently improves the performance of each foundation-model backbone, enabling all three to surpass the strongest baseline. This trend is consistent with the results reported on the benchmark datasets in the main paper.

\subsection{Class-level cell identity prediction performance}
\label{sec:additional-experiments-class-level}
\begin{table}[t]
\centering

\small
\setlength{\tabcolsep}{5pt}
\renewcommand{\arraystretch}{1.05}

\resizebox{\linewidth}{!}{
\begin{tabular}{lccccc}
\toprule
\textbf{Cell type}
& \textbf{Cells}
& \textbf{scTOP-rep on HVG + LR}
& \textbf{Geneformer}
& \textbf{CellRefine $\rightarrow$ FF}
& $\boldsymbol{\Delta}$ \textbf{vs. backbone} \\
\midrule

CD8+ Cytotoxic T
& 20,757
& 0.90
& 0.914
& 0.941
& +0.027 \\

CD8+/CD45RA+ Naive Cytotoxic
& 16,645
& 0.89
& 0.848
& 0.947
& +0.099 \\

CD56+ NK
& 8,775
& 0.94
& 0.896
& 0.965
& +0.069 \\

CD4+/CD25 T Reg
& 6,185
& 0.77
& 0.792
& 0.860
& +0.068 \\

CD19+ B
& 5,877
& 0.79
& 0.841
& 0.934
& +0.093 \\

CD4+/CD45RO+ Memory
& 3,059
& 0.70
& 0.669
& 0.800
& +0.131 \\

CD14+ Monocyte
& 2,847
& 0.86
& 0.845
& 0.924
& +0.079 \\

Dendritic
& 2,095
& 0.75
& 0.708
& 0.836
& +0.128 \\

CD4+/CD45RA+/CD25$^{-}$ Naive T
& 1,871
& 0.63
& 0.500
& 0.781
& +0.281 \\

CD34+
& 242
& 0.84
& 0.924
& 0.965
& +0.041 \\

CD4+ T Helper2
& 97
& 0.06
& 0.126
& 0.320
& +0.194 \\

\midrule
\textbf{Macro-F1}
&
& \textbf{0.739}
& \textbf{0.733}
& \textbf{0.843}
& \textbf{+0.110} \\

\bottomrule
\end{tabular}
}
\caption{Class-level cell identity prediction performance on the Blood dataset. {\OurModel} is applied to the Geneformer backbone. $\Delta$ denotes the improvement over the Geneformer backbone.}
\label{tab:blood_celltype_results}
\end{table}

Table~\ref{tab:blood_celltype_results} presents class-level cell identity prediction performance. We draw below observations.

(1) Post-pretraining improves every class, and CellRefine is the only method that dominates the classical baseline across the board. CellRefine raises F1 on all eleven cell types relative to the Geneformer backbone (macro 0.733→0.843, +15.0\%), and it exceeds the strong classical baseline on all eleven, whereas the off-the-shelf backbone does not (Geneformer, at 0.733, is below the 0.739 classical baseline).

(2) The gains are concentrated on the transcriptionally confusable subtypes, where the off-the-shelf model is weakest. The abundant, distinct types (CD8+ Cytotoxic, NK, CD34+) are already near-ceiling and gain modestly (+0.03 to +0.07). The large gains fall on the closely related subsets that all methods confuse: CD4+ Naive T (0.500→0.781, +56\%), CD4+ Memory (0.669→0.800), Dendritic (0.708→0.836), the CD8 naive/cytotoxic pair (0.848→0.947), and Treg (0.792→0.860). The CD4+ T Helper2 class improves (0.06→0.126→0.320) but still remains the hardest class.

\subsection{Experiments with additional parameter efficient baselines}
\label{sec:additional-experiments-baselines}
In addition to the baselines in section~\ref{sec:baselines}, we tested the following parameter-efficient variations of the strongest performing foundation model, Geneformer~\cite{theodoris2023transfer}, aiming to observe the learning efficiency. 

\textbf{Linear Probing (LP)}: freeze the pre-trained weights and train a linear layer with the task loss. \textbf{Last-Layer Fine-Tuning (LL)}: fine-tune only the final layer of the foundation model with the task loss. \textbf{Full Fine-Tuning (FF)}: fine-tune the entire foundation model with the task loss. \textbf{MLM LL $\rightarrow$ LL}: first perform last-layer MLM post-training with the given dataset and then perform last-layer fine-tuning with the task loss. \textbf{MLM $\rightarrow$ FF}: first perform continued MLM post-training with the given dataset and then perform full fine-tuning with the task loss. \textbf{LoRA}: perform parameter-efficient fine-tuning with a frozen backbone and trainable LoRA adapters of the foundational model \cite{hu2022lora}. \textbf{MLM LoRA $\rightarrow$ LoRA}: first perform continued MLM post-training updating only the LoRA adapter parameters, followed by LoRA fine-tuning with the task loss.

In addition to the baselines, we also define three variants of \text{\OurModel} for comparison. \textbf{\text{\OurModel} LL $\rightarrow$ LL}: perform post-pretraining with the given dataset using \text{\OurModel} but only update the last layer of the foundational model. \textbf{\text{\OurModel} LoRA $\rightarrow$ LoRA}: perform post-pretraining with the given dataset using \text{\OurModel} but use LoRA adapters. \textbf{\text{\OurModel} $\rightarrow$ FF}: perform post-pretraining with the given dataset using \text{\OurModel} and update the full foundational model.

\begin{table}[!t]
\centering
\small
\setlength{\tabcolsep}{5pt}

\resizebox{\textwidth}{!}{
\begin{tabular}{lccccccccccccc}
\toprule
\textbf{Method} & \textbf{Blood} & \textbf{Pancreas} & \textbf{Myeloid} & \textbf{Liver} & 
\textbf{LivST1} & \textbf{LivST2} & \textbf{MS} & \textbf{Lung} & \textbf{PBMC10K} & \textbf{Heart} & \textbf{Avg.} $\uparrow$ \\
\midrule
LP & 0.14 & 0.12 & 0.14 & 0.11 & 0.13 & 0.11 & 0.09 & 0.16 & 0.19 & 0.13 & 0.13 \\
LL & 0.54 & 0.49 & 0.21 & 0.16 & 0.37 & 0.28 & 0.41 & 0.48 & 0.62 & 0.49 & 0.41 \\
LoRA & 0.60 & 0.68 & 0.30 & 0.18 & 0.50 & 0.51 & 0.68 & 0.89 & 0.93 & 0.77 & 0.60 \\ 
FF & 0.67 & 0.69 & 0.34 & 0.27 & 0.64 & 0.60 & 0.70 & 0.92 & 0.93 & 0.86 & 0.66 \\
MLM LL $\rightarrow$ LL & 0.55 & 0.56 & 0.20 & 0.16 & 0.42 & 0.31 & 0.40 & 0.51 & 0.61 & 0.47 & 0.42 \\
MLM LoRA $\rightarrow$ LoRA & 0.63 & 0.72 & 0.32 & 0.18 & 0.51 & 0.53 & 0.65 & 0.90 & 0.92 & 0.8 & 0.62 \\
MLM $\rightarrow$ FF & 0.73 & 0.72 & 0.35 & 0.28 & 0.68 & 0.61 & 0.72 & 0.95 & 0.95 & 0.86 & 0.69 \\
\rowcolor{lightpurple}
\text{\OurModel} LL $\rightarrow$ LL (Ours) & 0.68 & 0.74 & 0.24 & 0.16 & 0.40 &  0.41 &  0.44 & 0.49 & 0.64 & 0.51 & 0.47 \\
\rowcolor{lightpurple}
\text{\OurModel} LoRA $\rightarrow$ LoRA (Ours) & 0.79 & 0.82 & 0.37 & 0.31 & 0.71 & 0.70 & 0.71 & 0.93 & 0.95 & 0.85 & 0.71 \\
\rowcolor{lightpurple}
\text{\OurModel} $\rightarrow$ FF (Ours) & \textbf{0.84} & \textbf{0.83} & \textbf{0.39} & \textbf{0.38} & \textbf{0.77} & \textbf{0.73} & \textbf{0.75} & \textbf{0.97} & \textbf{0.96} & \textbf{0.91} & \textbf{0.75} \\
\midrule
\end{tabular}
}
\caption{On-domain cell identity prediction performance across 10 benchmark datasets, measured by macro-F1 (higher is better). Variants of \text{\OurModel} are highlighted in purple, and the best overall result for each dataset is shown in bold.}
\label{Cell identity prediction appendix results}

\end{table}

\begin{table}
\centering
\small
\setlength{\tabcolsep}{4pt}
\begin{tabular}{lcccc}
\toprule
\textbf{Method}
& \multicolumn{2}{c}{\textbf{LiverST1}}
& \multicolumn{2}{c}{\textbf{LiverST2}} \\
& Cos $\uparrow$
& Cor $\uparrow$
& Cos $\uparrow$
& Cor $\uparrow$ \\
\midrule
LoRA
& 0.43 & 0.28
& 0.42 & 0.29 \\
FF
& 0.48 & 0.30
& 0.49 & 0.32 \\
\rowcolor{lightpurple}
\text{\OurModel} LoRA $\rightarrow$ LoRA (Ours)
& 0.46 & 0.29
& 0.46 & 0.31 \\
\rowcolor{lightpurple}
\text{\OurModel} $\rightarrow$ FF (Ours)
& \textbf{0.52} & \textbf{0.31}
& \textbf{0.51} & \textbf{0.34} \\
\bottomrule
\end{tabular}
\caption{
Spatial transcriptomics imputation performance on LivST1 and LivST2.
Best results are shown in bold.
}
\label{tab:spatial-transcriptomics-appendix-results}
\end{table}

\begin{table}[h]
\centering
\small
\setlength{\tabcolsep}{5pt}

\resizebox{\linewidth}{!}{%

\begin{tabular}{lccccccccccccc}
\toprule
\textbf{Method} & \textbf{CD4 T} & \textbf{CD8 T} & \textbf{NK} & \textbf{B} & 
\textbf{CD14+ Mono} & \textbf{FCGR3A+ Mono} & \textbf{Dendritic} & \textbf{Avg.} $\uparrow$ \\
\midrule

LoRA & 0.18 & 0.20 & 0.22 & 0.24 & 0.49 & 0.48 & \textbf{0.46} & 0.32 \\ 

FF & 0.20 & 0.23 & 0.23 & 0.24 & 0.52 & 0.49 & 0.44 & 0.34 \\

MLM LoRA $\rightarrow$ LoRA & 0.18 & 0.21 & 0.22 & 0.23 & 0.50 & 0.48 & 0.45 & 0.32 \\

MLM $\rightarrow$ FF & 0.20 & 0.23 & \textbf{0.24} & 0.25 & 0.53 & \textbf{0.50} & 0.45 & 0.34 \\

\rowcolor{lightpurple}
\text{\OurModel} LoRA $\rightarrow$ LoRA (Ours)
& \textbf{0.23} & 0.23 & 0.23 & 0.24 & \textbf{0.54} & 0.49 & \textbf{0.46} & 0.34 \\

\rowcolor{lightpurple}
\text{\OurModel} $\rightarrow$ FF (Ours)
& \textbf{0.23} & \textbf{0.25} & \textbf{0.24} & \textbf{0.26}
& \textbf{0.54} & \textbf{0.50} & \textbf{0.49} & \textbf{0.36} \\

\midrule
\end{tabular}
}
\caption{Zero-shot gene perturbation response prediction results. Variants of \text{\OurModel} are highlighted in purple, and the best overall result for each dataset is shown in bold.}
\label{tab:gene-pertubation-appendix}

\end{table}

In addition to Table~\ref{Cell identity prediction main results} in the main paper, here we report extended results on on-domain cell identity prediction results in Table~\ref{Cell identity prediction appendix results}, comparing \text{\OurModel} with the set of baselines across 10 datasets. \text{\OurModel} consistently achieves stronger performance in terms of macro-F1 across all datasets, outperforming existing alignment methods. Notable performance gains of over 15\% relative to the strongest baseline MLM $\rightarrow$ FF are observed on Blood, Pancreas, Liver, LivST1, and LivST2, which exhibit pronounced class imbalance and substantial cross-dataset generalization challenges. Overall, Full fine-tuning (FF) achieves an average macro-F1 of 0.66, while MLM alignment followed by full fine-tuning (MLM $\rightarrow$ FF) improves this to 0.69. In contrast, full fine-tuning with CellRefine (\text{\OurModel} $\rightarrow$ FF) achieves a substantially higher macro-F1 of 0.75. These improvements are consistent across fine-tuning strategies, including full and last-layer fine-tuning, and LoRA, which are strong evidence of the efficiency and effectiveness of \text{\OurModel}.

In the spatial transcriptomics imputation task, we observe a trend similar to that in the cell identity prediction task: \text{\OurModel} consistently outperforms the baselines. The results are presented in Table~\ref{tab:spatial-transcriptomics-appendix-results}. We note that, because the objective in this task is to impute masked genes, it is naturally formulated as a masked language modeling problem. As a result, the only applicable baselines are \text{FF} and \text{LoRA}, both of which are trained using the masked language modeling loss.

Table~\ref{tab:gene-pertubation-appendix} presents Pearson Correlation Coefficients of the zero-shot perturbation response prediction task. We note that when {\OurModel} post-pretraining is applied, LoRA variant outperforms the LoRA variant of Geneformer without {\OurModel} post-pretraining. 

\subsection{Experiments Isolating the Contribution of Marker Biology}
\label{sec:additional-experiments-contribution-marker-biology}

In this section, we present the results isolating the contribution of marker biology. We used Geneformer backbone for all the runs presented in Table~\ref{tab:cellrefine_ablation}, Table~\ref{tab:st_ablation}, and Table~\ref{tab:perturbation_ablation}. MLM $\rightarrow$ FF refers to the initial model prior to applying {\OurModel} post-pretraining, and {\OurModel} $\rightarrow$ FF refers to the model after applying {\OurModel} post-pretraining.

\begin{table}[t]
\centering

\small
\setlength{\tabcolsep}{5pt}
\renewcommand{\arraystretch}{1.05}

\resizebox{\linewidth}{!}{
\begin{tabular}{lcccc}
\toprule
\textbf{Method}
& \textbf{Blood}
& \textbf{Liver}
& \textbf{Myeloid}
& \textbf{Avg.} $\uparrow$ \\
\midrule

MLM $\rightarrow$ FF
& $0.731 \pm 0.008$
& $0.284 \pm 0.011$
& $0.354 \pm 0.006$
& $0.456 \pm 0.008$ \\

(1) Compute-matched MLM $\rightarrow$ FF
& $0.729 \pm 0.007$
& $0.291 \pm 0.006$
& $0.347 \pm 0.009$
& $0.456 \pm 0.007$ \\

(2) Supervised contrastive, no markers
& $0.722 \pm 0.010$
& $0.274 \pm 0.009$
& $0.350 \pm 0.008$
& $0.449 \pm 0.009$ \\

(3) Randomly initialized learnable prototype alignment $\rightarrow$ FF
& $0.751 \pm 0.004$
& $0.343 \pm 0.005$
& $0.366 \pm 0.005$
& $0.487 \pm 0.005$ \\

(4) Random marker alignment $\rightarrow$ FF
& $0.742 \pm 0.009$
& $0.334 \pm 0.007$
& $0.351 \pm 0.009$
& $0.476 \pm 0.008$ \\

(5) Shuffled marker alignment $\rightarrow$ FF
& $0.789 \pm 0.009$
& $0.345 \pm 0.008$
& $0.371 \pm 0.012$
& $0.502 \pm 0.010$ \\

(6) 25\% markers
& $0.745 \pm 0.007$
& $0.362 \pm 0.007$
& $0.363 \pm 0.003$
& $0.490 \pm 0.006$ \\

(7) 50\% markers
& $0.773 \pm 0.006$
& $\mathbf{0.382 \pm 0.004}$
& $0.364 \pm 0.003$
& $0.506 \pm 0.004$ \\

(8) 75\% markers
& $0.820 \pm 0.004$
& $0.380 \pm 0.004$
& $0.381 \pm 0.007$
& $0.527 \pm 0.005$ \\

\rowcolor{lightpurple}
\OurModel $\rightarrow$ FF
& $\mathbf{0.843 \pm 0.005}$
& $\mathbf{0.382 \pm 0.008}$
& $\mathbf{0.391 \pm 0.006}$
& $\mathbf{0.539 \pm 0.006}$ \\

\bottomrule
\end{tabular}
}
\caption{Ablation study isolating the contribution of marker biology in cell identity prediction. Results are reported as mean $\pm$ standard deviation in macro-F1, where higher is better.}
\label{tab:cellrefine_ablation}
\end{table}

In the cell identity prediction task presented in Table~\ref{tab:cellrefine_ablation}, the compute-matched MLM $\rightarrow$ FF performs the same as the baseline. Supervised contrastive learning without markers is almost on par with the baseline, indicating that generic label supervision alone does not help. The content-free prototype controls, namely randomly initialized learnable prototypes and random-marker alignment, remain well below our method. As we increase the fraction of marker genes, macro-F1 rises gradually, with {\OurModel} (full markers) obtaining the best result of 0.539.

\begin{table}[t]
\centering

\small
\setlength{\tabcolsep}{5pt}
\renewcommand{\arraystretch}{1.05}

\resizebox{\linewidth}{!}{
\begin{tabular}{lccc}
\toprule
\textbf{Method}
& \textbf{LiverST1}
& \textbf{LiverST2}
& \textbf{Avg.} $\uparrow$ \\
\midrule

MLM $\rightarrow$ FF
& $0.477 \pm 0.011$
& $0.491 \pm 0.012$
& $0.484 \pm 0.012$ \\

(1) Compute-matched MLM $\rightarrow$ FF
& $0.492 \pm 0.008$
& $0.491 \pm 0.007$
& $0.492 \pm 0.008$ \\

(2) Supervised contrastive, no markers
& $0.475 \pm 0.008$
& $0.476 \pm 0.013$
& $0.476 \pm 0.011$ \\

(3) Randomly initialized learnable prototype alignment $\rightarrow$ FF
& $0.490 \pm 0.012$
& $0.486 \pm 0.008$
& $0.488 \pm 0.010$ \\

(4) Random marker alignment $\rightarrow$ FF
& $0.481 \pm 0.009$
& $0.483 \pm 0.011$
& $0.482 \pm 0.010$ \\

(5) 25\% markers
& $0.492 \pm 0.011$
& $0.494 \pm 0.008$
& $0.493 \pm 0.010$ \\

(6) 50\% markers
& $0.496 \pm 0.007$
& $0.501 \pm 0.009$
& $0.499 \pm 0.008$ \\

(7) 75\% markers
& $0.514 \pm 0.008$
& $0.508 \pm 0.014$
& $0.511 \pm 0.011$ \\

\rowcolor{lightpurple}
\OurModel $\rightarrow$ FF
& $\mathbf{0.526 \pm 0.013}$
& $\mathbf{0.514 \pm 0.008}$
& $\mathbf{0.520 \pm 0.011}$ \\

\bottomrule
\end{tabular}
}
\caption{Ablation isolating the contribution of marker biology for spatial transcriptomics imputation. Results are reported as mean $\pm$ standard deviation in cosine similarity, where higher is better.}
\label{tab:st_ablation}
\end{table}

In the spatial transcriptomics imputation task presented in Table~\ref{tab:st_ablation}, we observe the same behavior as in the previous task. The compute-matched and content-free controls stay near the baseline, while performance again increases with marker coverage and peaks for {\OurModel} (0.520).

\begin{table}[t]
\centering

\small
\setlength{\tabcolsep}{5pt}
\renewcommand{\arraystretch}{1.05}

\resizebox{\linewidth}{!}{
\begin{tabular}{lccc}
\toprule
\textbf{Method}
& \textbf{CD4 T}
& \textbf{CD8 T}
& \textbf{Avg.} $\uparrow$ \\
\midrule

MLM $\rightarrow$ FF
& $0.204 \pm 0.002$
& $0.231 \pm 0.001$
& $0.218 \pm 0.001$ \\

(1) Compute-matched MLM $\rightarrow$ FF
& $0.208 \pm 0.0009$
& $0.225 \pm 0.002$
& $0.217 \pm 0.001$ \\

(2) Supervised contrastive, no markers
& $0.200 \pm 0.003$
& $0.233 \pm 0.002$
& $0.217 \pm 0.002$ \\

(3) Randomly initialized learnable prototype alignment $\rightarrow$ FF
& $0.211 \pm 0.004$
& $0.231 \pm 0.001$
& $0.221 \pm 0.002$ \\

(4) Random markers
& $0.205 \pm 0.001$
& $0.233 \pm 0.005$
& $0.219 \pm 0.003$ \\

(5) 25\% markers
& $0.207 \pm 0.003$
& $0.230 \pm 0.007$
& $0.219 \pm 0.005$ \\

(6) 50\% markers
& $0.218 \pm 0.0009$
& $0.240 \pm 0.005$
& $0.229 \pm 0.003$ \\

(7) 75\% markers
& $\mathbf{0.232 \pm 0.001}$
& $0.244 \pm 0.003$
& $0.238 \pm 0.002$ \\

\rowcolor{lightpurple}
\OurModel $\rightarrow$ FF
& $\mathbf{0.232 \pm 0.002}$
& $\mathbf{0.245 \pm 0.001}$
& $\mathbf{0.239 \pm 0.002}$ \\

\bottomrule
\end{tabular}
}
\caption{Ablation study isolating the contribution of marker biology for zero-shot gene perturbation response prediction. Results are reported as mean $\pm$ standard deviation in Pearson correlation coefficient, where higher is better.}
\label{tab:perturbation_ablation}
\end{table}

In the zero-shot gene perturbation response prediction task presented in Table~\ref{tab:perturbation_ablation}, the same observation continues. The compute-matched and content-free controls remain at 0.21–0.22, the marker-coverage ladder rises, and {\OurModel} attains 0.239. Although the gains here are small relative to the other two tasks, the low run-to-run variance means the improvement of {\OurModel} over the baseline and the content-free controls is nonetheless distinguishable from noise, and the direction of the trend is consistent with the other two tasks.

\subsection{Loss Function Hyperparameter Sensitivity Analysis}
\label{sec:loss-hp}

\begin{table}[t]
\centering

\small
\setlength{\tabcolsep}{6pt}
\renewcommand{\arraystretch}{1.05}

\begin{tabular}{lcc}
\toprule
$\boldsymbol{\lambda_1}$
& \textbf{Liver}
& \textbf{Myeloid} \\
\midrule

$0$
& $0.321 \pm 0.012$
& $0.366 \pm 0.006$ \\

$0.1 \times 0.5$
& $0.320 \pm 0.009$
& $0.366 \pm 0.008$ \\

$0.3 \times 0.5$
& $0.323 \pm 0.014$
& $0.367 \pm 0.008$ \\

$\mathbf{1 \times 0.5}$
& $\mathbf{0.382 \pm 0.008}$
& $\mathbf{0.391 \pm 0.006}$ \\

$3 \times 0.5$
& $0.362 \pm 0.010$
& $0.387 \pm 0.006$ \\

$10 \times 0.5$
& $0.341 \pm 0.009$
& $0.362 \pm 0.009$ \\

\bottomrule
\end{tabular}
\caption{Sensitivity analysis of $\lambda_1$, the prototype loss coefficient. Results are reported as mean $\pm$ standard deviation.}
\label{tab:lambda1_sensitivity}
\end{table}

\begin{table}[t]
\centering

\small
\setlength{\tabcolsep}{6pt}
\renewcommand{\arraystretch}{1.05}

\begin{tabular}{lcc}
\toprule
$\boldsymbol{\lambda_2}$
& \textbf{Liver}
& \textbf{Myeloid} \\
\midrule

$0$
& $0.363 \pm 0.011$
& $0.370 \pm 0.010$ \\

$0.1 \times 1$
& $0.371 \pm 0.008$
& $0.370 \pm 0.009$ \\

$0.3 \times 1$
& $0.375 \pm 0.009$
& $0.373 \pm 0.006$ \\

$\mathbf{1 \times 1}$
& $\mathbf{0.382 \pm 0.008}$
& $\mathbf{0.391 \pm 0.006}$ \\

$3 \times 1$
& $0.380 \pm 0.010$
& $0.390 \pm 0.009$ \\

$10 \times 1$
& $0.381 \pm 0.009$
& $0.390 \pm 0.011$ \\

\bottomrule
\end{tabular}
\caption{Sensitivity analysis of $\lambda_2$, the lineage loss coefficient. Results are reported as mean $\pm$ standard deviation.}
\label{tab:lambda2_sensitivity}
\end{table}

\begin{table}[t]
\centering

\small
\setlength{\tabcolsep}{6pt}
\renewcommand{\arraystretch}{1.05}

\begin{tabular}{lcc}
\toprule
$\boldsymbol{\lambda_3}$
& \textbf{Liver}
& \textbf{Myeloid} \\
\midrule

$0$
& $0.361 \pm 0.006$
& $0.372 \pm 0.009$ \\

$0.1 \times 0.03$
& $0.361 \pm 0.013$
& $0.372 \pm 0.012$ \\

$0.3 \times 0.03$
& $0.364 \pm 0.014$
& $0.373 \pm 0.009$ \\

$\mathbf{1 \times 0.03}$
& $\mathbf{0.382 \pm 0.008}$
& $\mathbf{0.391 \pm 0.006}$ \\

$3 \times 0.03$
& $0.303 \pm 0.010$
& $0.351 \pm 0.009$ \\

$10 \times 0.03$
& $0.276 \pm 0.020$
& $0.348 \pm 0.009$ \\

\bottomrule
\end{tabular}
\caption{Sensitivity analysis of $\lambda_3$, the GMVE loss coefficient. Results are reported as mean $\pm$ standard deviation.}
\label{tab:lambda3_sensitivity}
\end{table}

For selecting values for $\boldsymbol{\lambda_1}$, $\boldsymbol{\lambda_2}$, and $\boldsymbol{\lambda_3}$, we initialized the loss weights heuristically to balance the magnitudes of the four loss terms, then refined them with a coarse grid search on a held-out validation set, yielding $\boldsymbol{\lambda_1}$ = 0.5, $\boldsymbol{\lambda_2}$ = 1, and $\boldsymbol{\lambda_3}$ = 0.03. We performed a sensitivity analysis on the two most batch-affected datasets, Liver and Myeloid, where the regularization terms are expected to matter most, and sensitivity is therefore most informative. We use a one-at-a-time protocol: each coefficient is varied over a multiplicative grid — {0, 0.1×, 0.3×, 1×, 3×, 10×} of its selected value, while the other two are held fixed at their selected values. In every table, the 1× row corresponds to the configuration used throughout the paper and is common to all three tables, while the 0 row removes that loss term (i.e., the corresponding component ablation). All entries are macro-F1 (mean ± s.d. over three runs).

Results of loss parameter sensitivity analysis for cell identity prediction are presented in Table~\ref{tab:lambda1_sensitivity}, Table~\ref{tab:lambda2_sensitivity}, and Table~\ref{tab:lambda3_sensitivity}. Across all three coefficients, the selected value (1×) attains the best macro-F1 on both datasets. The lineage weight $\boldsymbol{\lambda_2}$ is robust: performance is flat to within run-to-run variation across the full 0.1×–10× range \cite{weerasekara2022trends,weerasekara2024reinforcement}, because its margin-based separation loss vanishes once related subtypes are separated and therefore has no effect once the margin is met. For the prototype weight $\boldsymbol{\lambda_1}$, below the selected value the marker signal is too weak to reshape the embedding, and performance falls to the ablation level (Liver $\approx 0.32$). The GMVE weight $\boldsymbol{\lambda_3}$ is stable up to the selected value but degrades sharply when over-weighted (Liver 0.382→0.276 at 1×→10×), consistent with a KL term that induces posterior collapse in the post-pretraining stage.

\subsection{Compute overhead comparison}
\label{sec:compute-overhead}

{\OurModel} incurs a one-time cost relative to the initial baseline foundation model, which performs continued MLM/AR post-training in place of {\OurModel} post-pretraining. In Table~\ref{tab:compute_comparison}, we report training times for full finetuning Geneformer and Geneformer + {\OurModel}, for the cell identity prediction task (secs) (post-pretraining + fine-tuning) on three datasets. Both methods are trained to convergence under an identical early-stopping criterion. So the reported numbers are each method's genuine time-to-convergence rather than a fixed compute budget. 

\begin{table}[t]
\centering

\small
\setlength{\tabcolsep}{6pt}
\renewcommand{\arraystretch}{1.05}

\begin{tabular}{lccc}
\toprule
\textbf{Method}
& \textbf{Blood}
& \textbf{Liver}
& \textbf{Myeloid} \\
\midrule

Geneformer
& $18124.0813$
& $5072.1429$
& $15309.3107$ \\

\rowcolor{lightpurple}
Geneformer + {\OurModel}
& $22039.2802$
& $7200.2811$
& $19322.2311$ \\

\bottomrule
\end{tabular}
\caption{Computational overhead comparison between the initial foundation model and foundation model with {\OurModel} post-pretraining.}
\label{tab:compute_comparison}
\end{table}

The overhead is one-time, and to place it in context, it is incurred once per dataset during post-pretraining; it does not recur, and it does not affect deployment. The fine-tuned model has the same architecture and size as the backbone, so {\OurModel} adds nothing to inference latency or memory. Part of this cost is also recovered downstream: by structuring the latent space beforehand, the post-pretraining stage provides a better-initialized embedding, so the subsequent fine-tuning converges faster than fine-tuning from the MLM checkpoint, partially offsetting the added cost.

\subsection{Latent space qualitative analysis}
\label{subsec:latent-space}

\begin{figure}[h]
    \centering
    \label{umaps}
    \begin{subfigure}[t]{0.48\textwidth}
        \includegraphics[width=\textwidth]{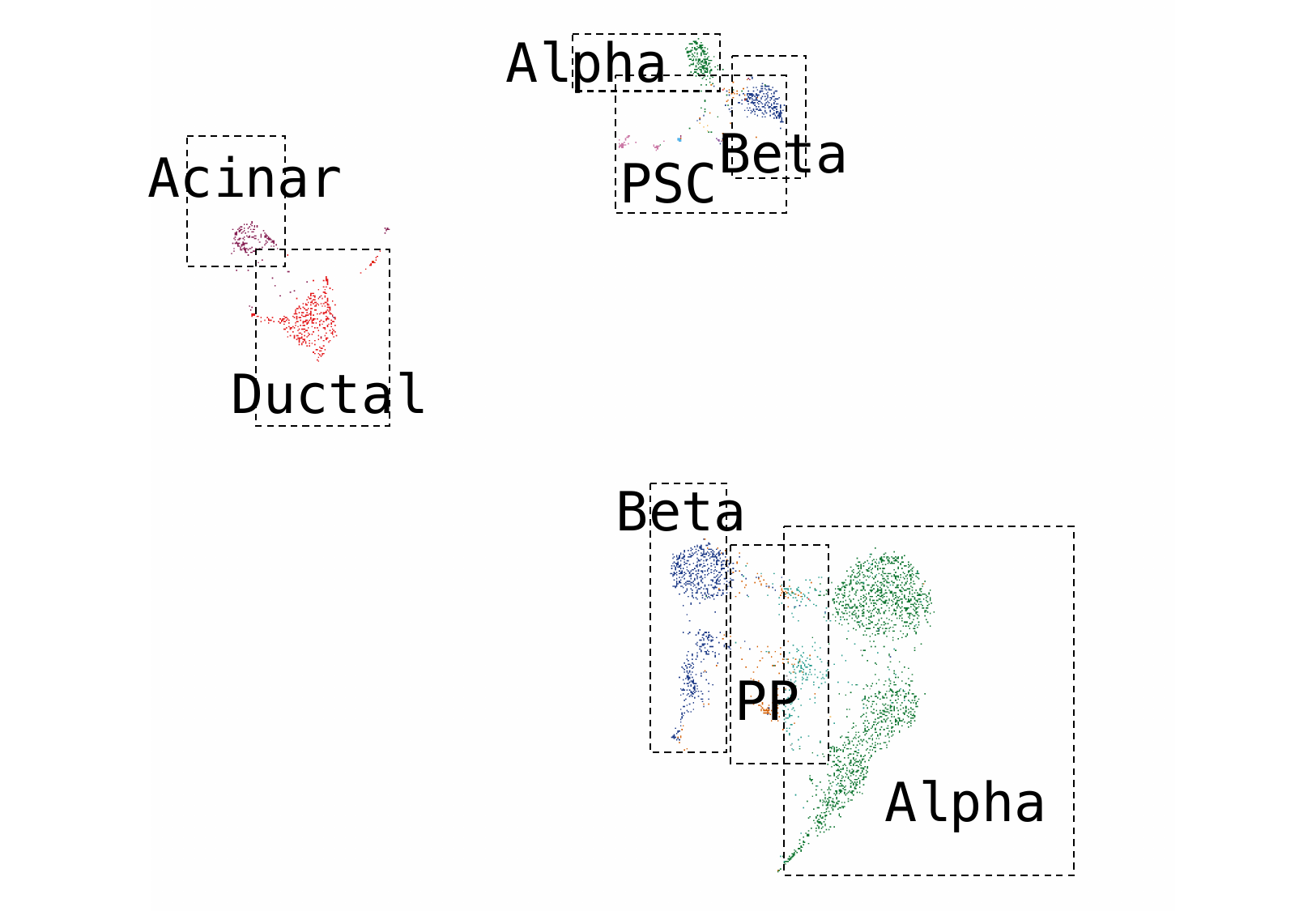}
        \caption{UMAP visualization of cell embeddings of the Pancreas dataset embeddings after applying MLM, before applying \text{\OurModel}.}
        \label{fig:umap2a}

    \end{subfigure}
    \hfill
    \begin{subfigure}[t]{0.48\textwidth}
        \includegraphics[width=\textwidth]{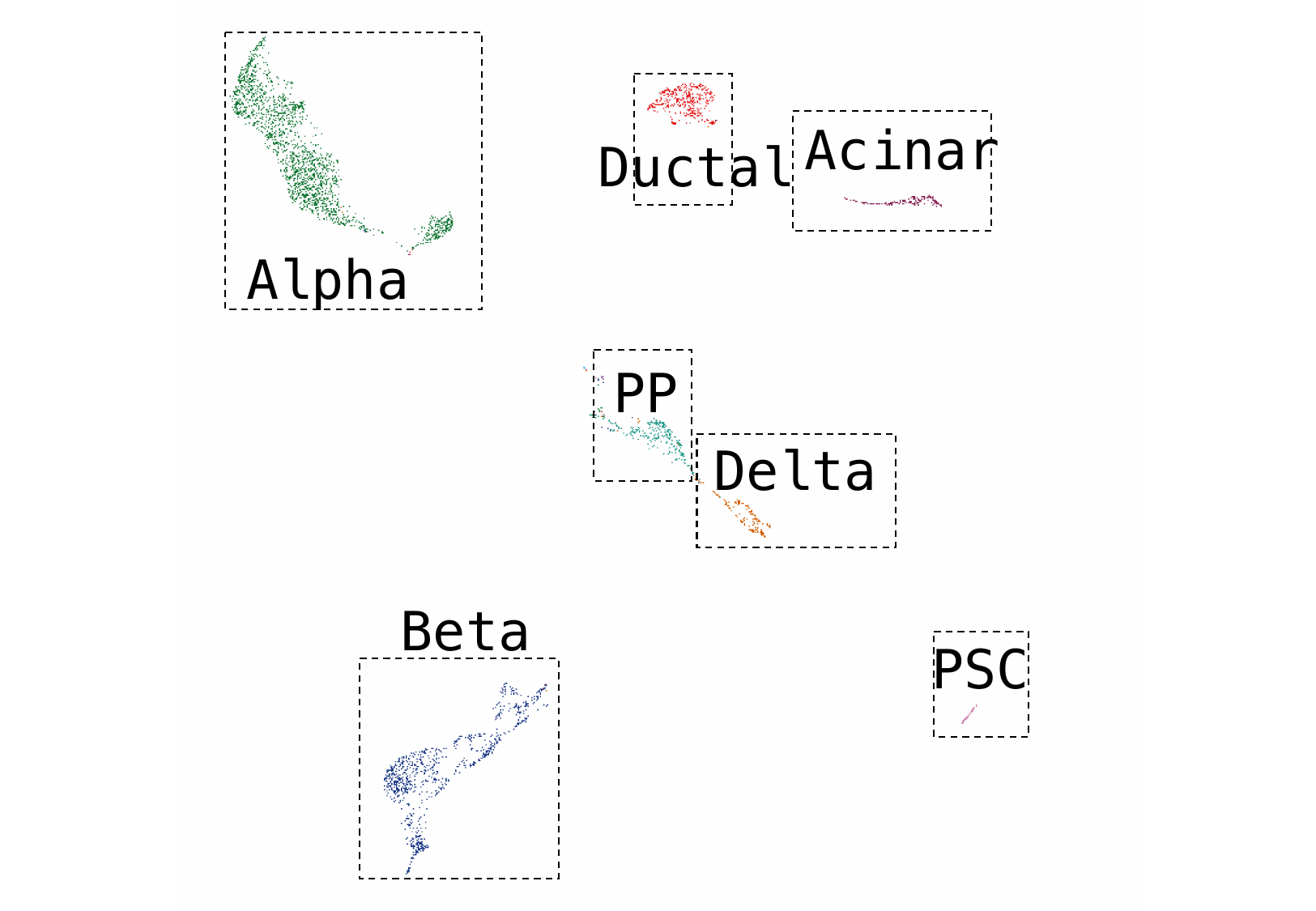}
        \caption{UMAP visualization of cell embeddings of the Pancreas dataset after post-pretraining with \text{\OurModel}.}
        \label{fig:umap2b}

    \end{subfigure}
    \caption{Visualization of latent embeddings for cells in the Pancreas dataset~\cite{chen2023transformer} before (Figure~\ref{fig:umap2a}) and after (Figure~\ref{fig:umap2b}) post-pretraining with \text{\OurModel}. Both instances correspond to the alignment stage prior to finetuning. Regions corresponding to cell clusters are outlined with dashed lines.}
    \label{additional-umaps-pancreas}
\end{figure}

Figure~\ref{additional-umaps-pancreas} illustrates the effect of refinement on the structure of cell embeddings using UMAP visualization. Prior to refinement (Figure~\ref{fig:umap2a}), embeddings learned via masked language modeling (MLM) capture coarse-grained structure, but closely related cell types—particularly niche or rare populations such as PP and certain endocrine subtypes—remain poorly separated, with substantial overlap between clusters. After applying \text{\OurModel} (Figure~\ref{fig:umap2b}), the embedding space exhibits markedly improved organization, with clearer cluster boundaries and enhanced separation of fine-grained cell types. Notably, niche cell populations that were previously entangled become more distinctly localized, suggesting that the refinement process improves the model’s ability to resolve subtle biological differences and enhances discriminability among closely related cell states.

\begin{figure}[h]
    \centering
    \begin{subfigure}[t]{0.48\textwidth}
        \includegraphics[width=\textwidth]{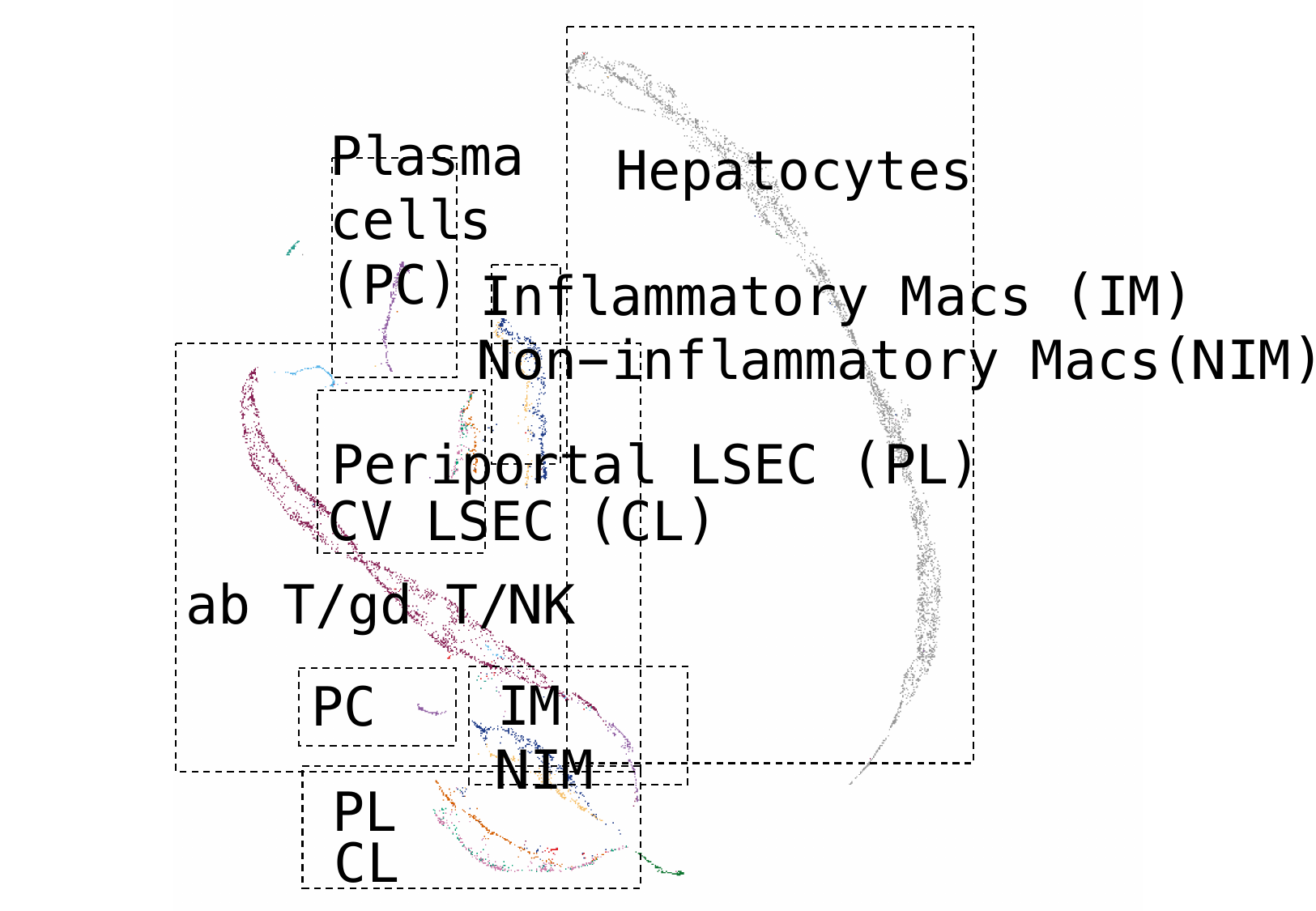}
        \caption{UMAP visualization of cell clusters of liver dataset after applying MLM LoRA.}
        \label{fig:umap3a}

    \end{subfigure}
    \hfill
    \begin{subfigure}[t]{0.48\textwidth}
        \includegraphics[width=\textwidth]{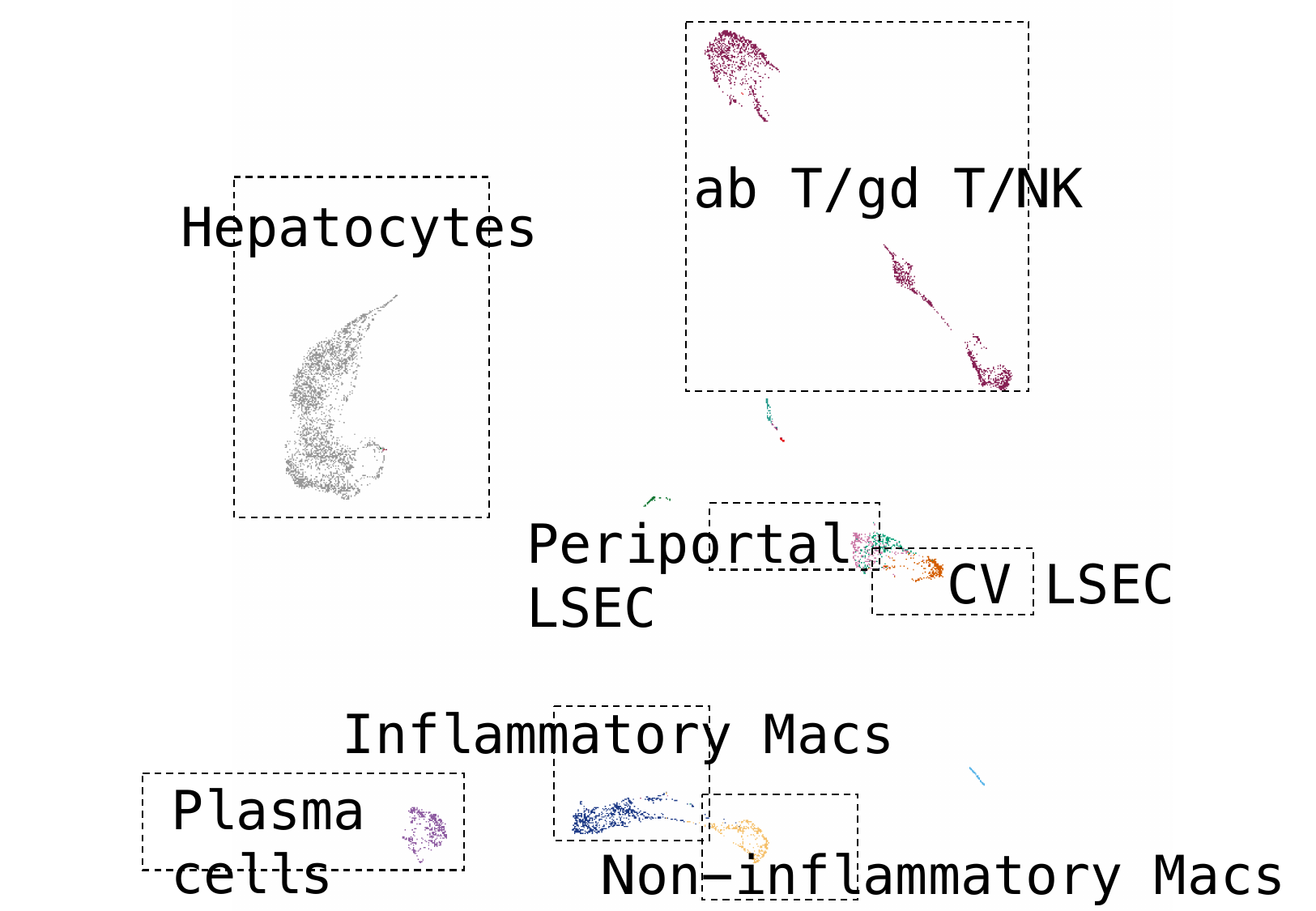}
        \caption{UMAP visualization of cells in the liver dataset after post-pretraining with \text{\OurModel} LoRA.}
        \label{fig:umap3b}

    \end{subfigure}
    \caption{Visualization of latent embeddings for cells in the liver dataset~\cite{lin2020scclassify} before (Figure~\ref{fig:umap3a}) and after (Figure~\ref{fig:umap3b}) post-pretraining with \text{\OurModel} LoRA. Both instances correspond to the alignment stage prior to finetuning. Regions corresponding to cell clusters are outlined with dashed lines.}
    \label{additional-umaps-liver}
\end{figure}

Figure~\ref{additional-umaps-liver} shows the UMAP visualization of cell embeddings for the liver dataset before and after post-training with \OurModel LoRA. Prior to refinement (Figure~\ref{fig:umap3a}), embeddings obtained after MLM with LoRA capture coarse cellular structure, but several related cell populations—such as inflammatory and non-inflammatory macrophages, LSEC subtypes (periportal and central vein), and lymphocyte populations—exhibit significant overlap and diffuse boundaries. After applying \text{\OurModel} LoRA (Figure~\ref{fig:umap3b}), the embedding space becomes more structured, with improved separation between biologically distinct cell types \cite{weerasekara2025cellclique}. In particular, closely related macrophage subpopulations and endothelial subtypes become more clearly delineated, while lymphocyte clusters (e.g., T/NK cells) are more compact and distinct. These results indicate that lightweight post-training via LoRA is sufficient to enhance the representation quality, enabling better resolution of fine-grained cellular heterogeneity in the liver dataset.

\begin{figure}[h]
    \centering
    \label{umaps}
    \begin{subfigure}[t]{0.48\textwidth}
        \includegraphics[width=\textwidth]{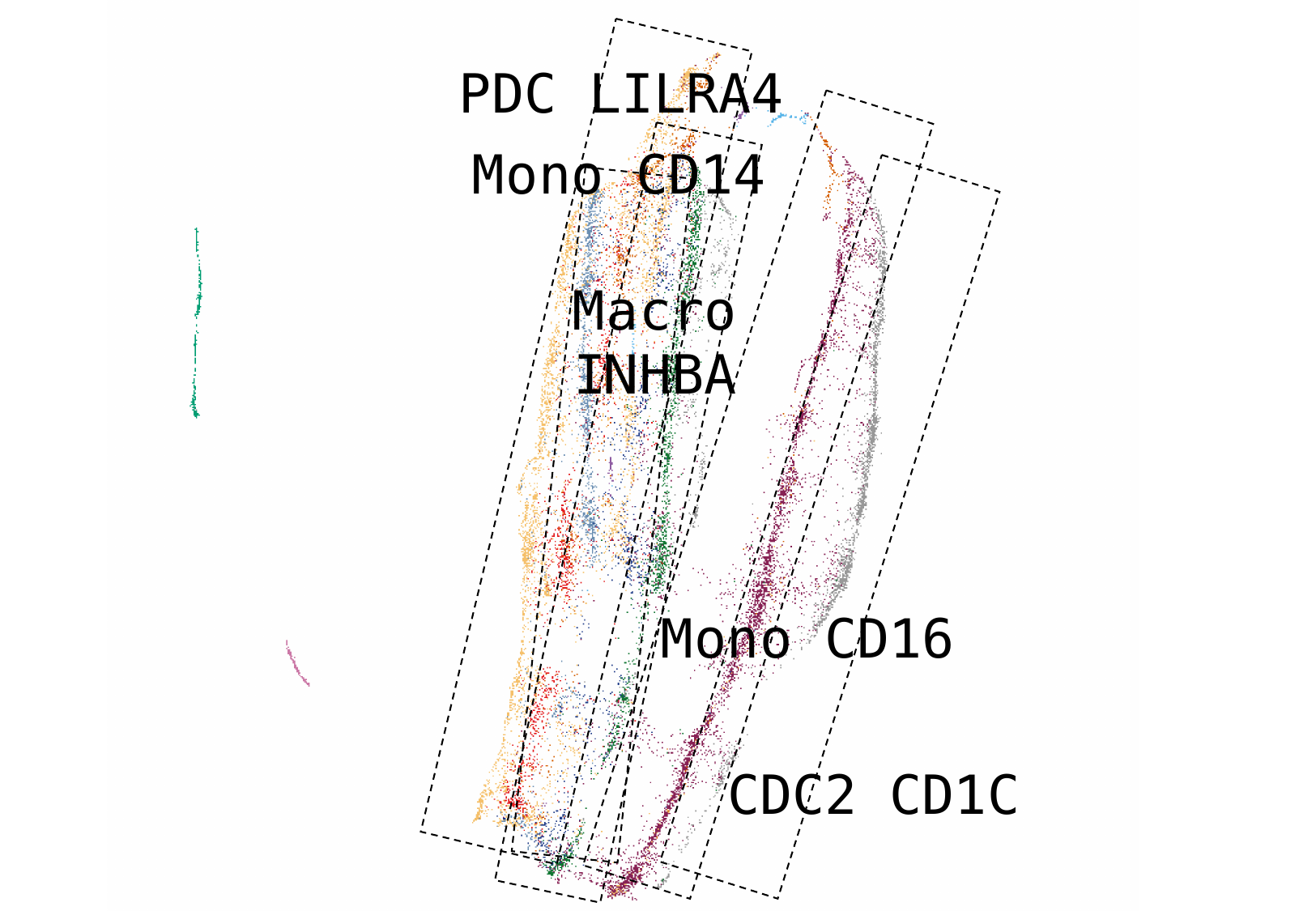}
        \caption{UMAP visualization of cell embeddings of the Myeloid dataset embeddings after applying MLM, before applying \text{\OurModel}.}
        \label{fig:umap2a}

    \end{subfigure}
    \hfill
    \begin{subfigure}[t]{0.48\textwidth}
        \includegraphics[width=\textwidth]{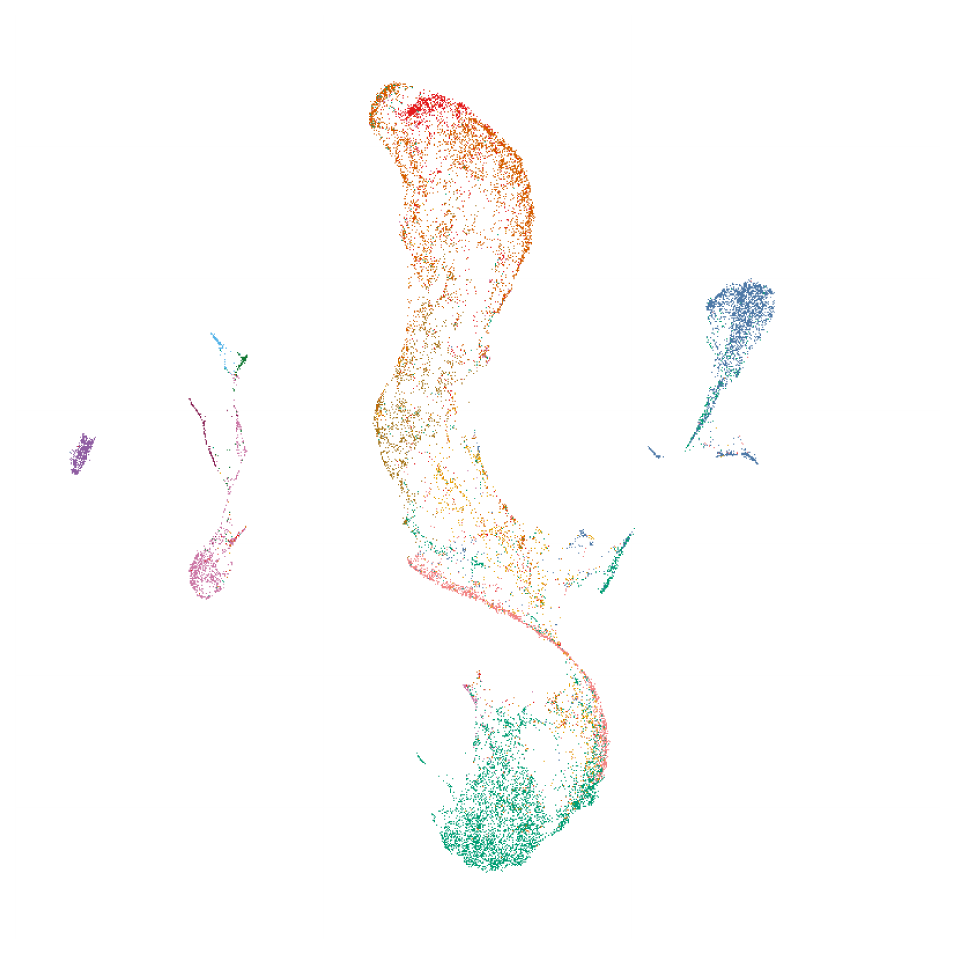}
        \caption{UMAP visualization of cell embeddings of the Myeloid dataset after post-pretraining with \text{\OurModel}.}
        \label{fig:umap2b}

    \end{subfigure}
    \caption{Visualization of latent embeddings for cells in the Myeloid dataset before (Figure~\ref{fig:umap2a}) and after (Figure~\ref{fig:umap2b}) post-pretraining with \text{\OurModel}. Both instances correspond to the alignment stage prior to finetuning. Regions corresponding to cell clusters are outlined with dashed lines.}
    \label{additional-umaps-myeloid}
\end{figure}

\begin{figure}[h]
    \centering
    
    \begin{subfigure}[t]{0.48\textwidth}
        \includegraphics[width=\textwidth]{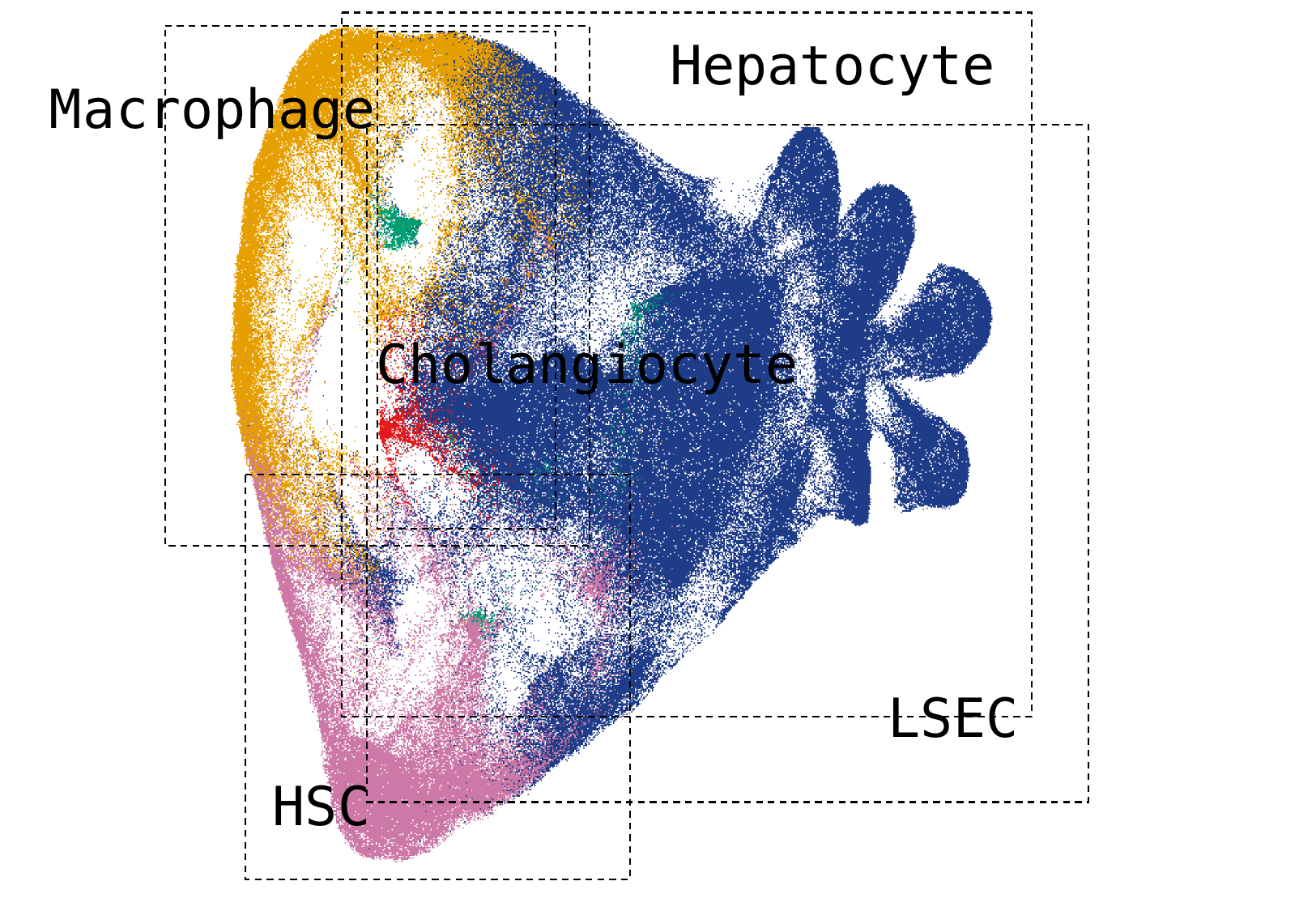}
        \caption{UMAP visualization of cell embeddings of the LivST2 dataset embeddings after aligning with MLM and finetuning for cell identity prediction task.}
        \label{fig:umap4a}

    \end{subfigure}
    \hfill
    \begin{subfigure}[t]{0.48\textwidth}
        \includegraphics[width=\textwidth]{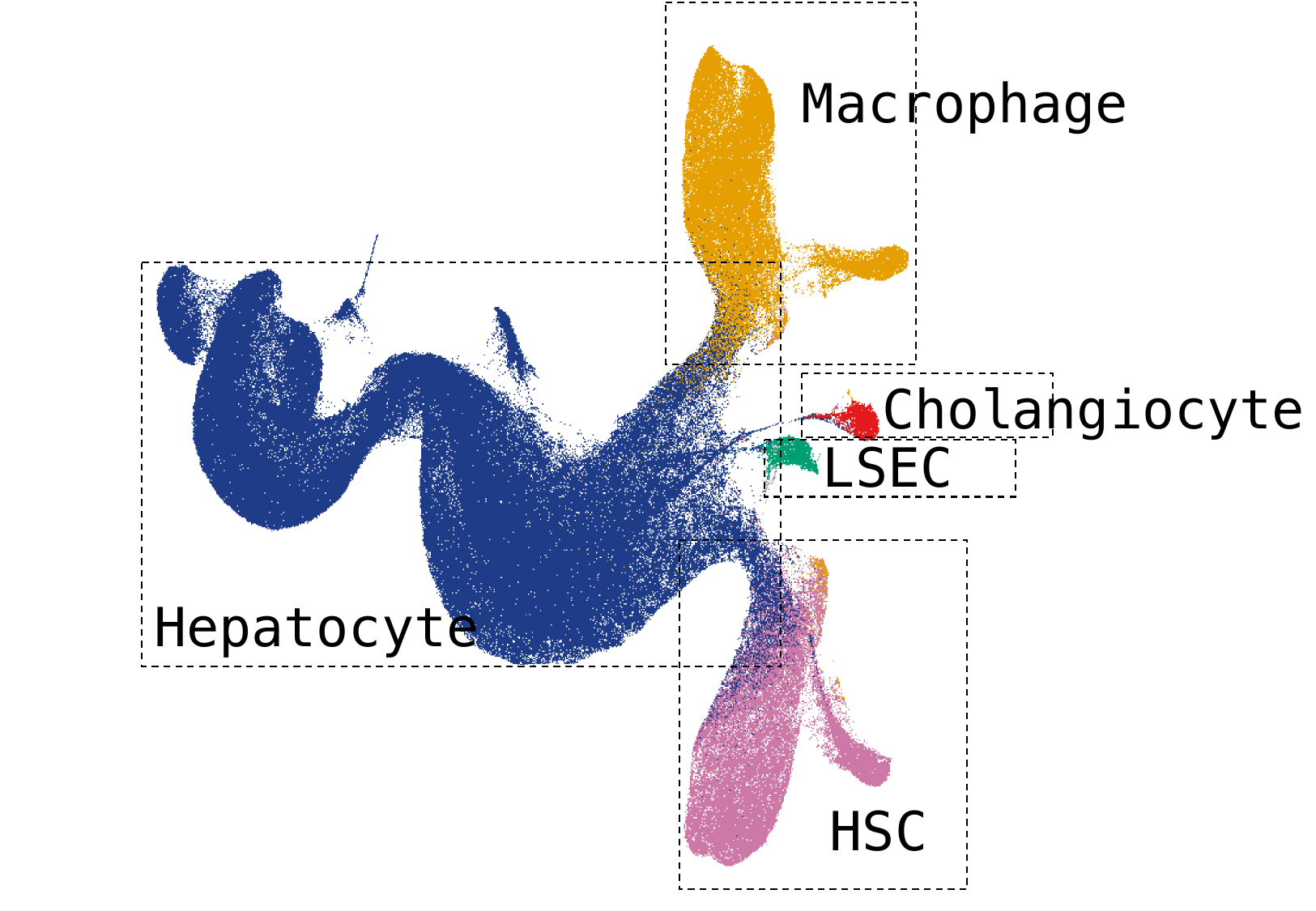}
        \caption{UMAP visualization of cell embeddings of the LivST2 dataset after post-pretraining with \text{\OurModel} and finetuning for cell itentity prediction task.}
        \label{fig:umap4b}

    \end{subfigure}
    \caption{Visualization of latent embeddings for tail cells in the LivST2 dataset before (Figure~\ref{fig:umap4a}) and after (Figure~\ref{fig:umap4b}) applying \text{\OurModel}, and after finetuning for cell identity prediction task.}
    \label{additional-umaps-livst2}
\end{figure}

Figure~\ref{additional-umaps-livst2} presents the UMAP visualization of cell embeddings for the LivST2 dataset before and after applying \text{\OurModel}, followed by fine-tuning for cell identity prediction. Prior to refinement (Figure~\ref{fig:umap4a}), the embedding space captures broad cellular structure, with major populations such as hepatocytes, macrophages, cholangiocytes, LSECs, and hepatic stellate cells (HSCs) partially organized \cite{weerasekara2026refineppo,weerasekara2025improvements}. However, several of these cell types exhibit overlapping regions and blurred boundaries, particularly among closely related epithelial and endothelial populations. After applying \text{\OurModel} (Figure~\ref{fig:umap4b}), the embeddings become substantially more structured, with clearer separation between distinct cell types. Notably, hepatocytes form a well-defined cluster, while smaller and more specialized populations such as cholangiocytes and LSECs are more distinctly localized. Additionally, macrophages and HSCs become more compact and separable from neighboring clusters. These results demonstrate that CellRefine improves the representation of spatial transcriptomics data, enhancing the model’s ability to resolve both major and niche cell populations during downstream cell identity prediction.

\section{Experimental Settings}
\label{sec:experimental-settings}

\begin{table}[h]
\centering
\small
\setlength{\tabcolsep}{8pt}

\begin{tabular}{lll}
\toprule
 & \textbf{Hyperparameter} & \textbf{Value} \\
\midrule

\multirow[t]{8}{*}{Cell Encoder ($f$)}
& Vocab size & 20275 \\
& Hidden size & 768 \\
& Number of hidden layers & 12 \\
& Max sequence length & 4096 \\
& Number of attention heads & 12 \\
& Dropout & 0.02 \\
& Activation & ReLU \\
& LayerNorm eps & 1e-12 \\

\midrule

\multirow[t]{8}{*}{Marker Encoder ($f_{Marker}$)}
& Vocab size & 20275 \\
& Hidden size & 768 \\
& Number of hidden layers & 2 \\
& Max sequence length & 200 \\
& Number of attention heads & 12 \\
& Dropout & 0.01 \\
& Activation & ReLU \\
& LayerNorm eps & 1e-12 \\

\midrule

\multirow[t]{6}{*}{LoRA Configuration}
& Rank ($r$) & 8 \\
& Alpha ($\alpha$) & 16 \\
& Dropout & 0.05 \\
& Target modules & \makecell[l]{query, key, value \\ (attention layers)} \\
& Bias & none \\
& Task type & encoder-only \\ 

\midrule

\multirow[t]{6}{*}{Post-pretraining}
& Optimizer & AdamW \\
& Scheduler & Linear \\
& Maximum learning rate & 1e-5 \\
& Warmup steps & 1,000\\
& Weight decay & 1e-3 \\
& Batch size & 12 \\

\bottomrule
\end{tabular}
\caption{Experiment configurations}
\label{Experiment configurations}
\end{table}

We perform the processing settings below across all datasets and training tasks.

\paragraph{General settings}
\begin{itemize}
    \item Perform quality control on all datasets, removing unrecognized cell types such as 'other', 'unknown' as well as cells with too few captured genes.
    \item For all tasks, we perform three random iterations and report the mean.
    \item All tasks were trained with early stopping based on validation loss, with a patience of 3 epochs.
    \item For MLM tasks, all models are trained until MLM loss reaches 3, following Geneformer \cite{theodoris2023transfer} pretraining. 
    \item Blood cell dataset \cite{zheng2017massively} uses a 80:20 train:test split and all other datasets use separate datasets for training and testing.

\end{itemize}

\paragraph{Training} All training was conducted in PyTorch v2.6.0 (CUDA 12.4 support), leveraging the Hugging Face Transformers version 4.44.2. We optimized model parameters using AdamW with a linear warmup phase of 1,000 steps to a peak learning rate of 1e-5, after which the learning rate was linearly decayed. Weight decay was set to 0.001. We ran the experiments on AMD EPYC 7702 64-core CPUs with 4 NVIDIA A100 GPUs with 80GB VRAM. Hyperparameter settings for the cell encoder, marker encoder, LoRA adapters, and post-pretraining are included in the Table ~\ref{Experiment configurations}

\section{Dataset details}
\label{sec:dataset-details}

\paragraph{\textbf{Blood}} The Blood dataset \cite{zheng2017massively} is a particularly challenging benchmark, comprising 68,450 peripheral blood mononuclear cells (PBMCs). It includes 11 closely related cell subtypes with subtle transcriptional differences, making accurate discrimination difficult. Additionally, the dataset exhibits a strongly imbalanced cell-type distribution, where certain populations are underrepresented. We use 80:20 train:test split of dataset for the cell identification task. 

\paragraph{\textbf{Pancreas}} The Pancreas dataset contains scRNA-seq data from five independent studies of human pancreas cells, reprocessed by \cite{chen2023transformer}. Following scGPT \cite{cui2024scgpt}, we split the five datasets to two reference datasets and three query datasets. The reference set contains 10,600 cells of 13 cell types, and the query set contains 4218 cells of 11 cell types. We used this dataset for the cell identification task. 

\paragraph{\textbf{Liver}} The Liver dataset sourced from \cite{lin2020scclassify} is a combination of the macParland and the Aizarani datasets. We utilized the macParland dataset, comprising 14 cell types and 8444 cells, and the reference set, and the Aizarani dataset, which includes 9162 cells and seven cell types that are present in the macParland dataset as the query dataset. We used this dataset for the cell identification task. 

\paragraph{\textbf{LiverST1}} LiverST1 is a human liver spatial transcriptomics dataset sourced from 10x Genomics \cite{10xgenomics2023liver}. It contains 90,408 cells and five cell types. This dataset was used for cell identification and spatial transcriptomics imputation tasks. This dataset was used to report performance when the models were trained with LiverST2 dataset.

\paragraph{\textbf{LiverST2}} LiverST2 is a human liver spatial transcriptomics dataset sourced from 10x Genomics \cite{10xgenomics2023liver}, containing 309,808 cells and five same cell types included in LivST1 dataset. We used this dataset for cell identity prediction and spatial transcriptomics imputation tasks, and was used to report performance when the models were trained with LiverST1 dataset.

\paragraph{\textbf{Myeloid}} Myeloid dataset \cite{cheng2021pan} can be accessed from Gene Expression Omnibus (GEO) database using accession number GSE154763. While the dataset contains nine different cancer types, following \cite{cui2024scgpt}, we used six cancer types:  UCEC, PAAD, THCA, LYM, cDC2, and kidney in the train set and three cancer types: MYE, OV-FTC, and ESCA in the test set.

\paragraph{\textbf{MS}} The MS dataset \cite{schirmer2019neuronal} has nine healthy samples, and 12 MS samples are included in the dataset. Following scGPT \cite{cui2024scgpt}, we use the control samples as the reference set and MS samples as the query set, creating a setting to test on out-of-distribution data. The reference set has a cell count of 7,844, and the query set has a cell count of 13,468. 

\paragraph{\textbf{Lung}} The lung dataset \cite{kim2020single} consists of 14 lung adenocarcinoma samples, with 30,472 cells. The dataset is publicly available at (\href{https://www.weizmann.ac.il/sites/3CA/lung}{https://www.weizmann.ac.il/sites/3CA/lung}). We partitioned it into a reference set of ten samples and a query set of four samples. The resulting reference set contains 24,746 cells, and the query set contains 7,747 cells.

\section{Evaluation Metrics}
\label{sec:evaluation-metrics}

\subsection{Cell Type Annotation}
\label{subsec:cell-type-annotation}

To evaluate performance on the cell type annotation task, we consider a multi-class classification setting with $K$ cell types. Given that several datasets are highly imbalanced, we primarily report the \textbf{macro F1-score}, which assigns equal importance to each class regardless of its frequency.

Let $y_i$ and $\hat{y}_i$ denote the ground-truth and predicted labels for sample $i$, respectively. For each class $k \in \{1, \dots, K\}$, we define true positives ($TP_k$), false positives ($FP_k$), and false negatives ($FN_k$).

We first compute class-wise precision and recall:
\begin{equation}
\text{Precision}_k = \frac{TP_k}{TP_k + FP_k}, \quad
\text{Recall}_k = \frac{TP_k}{TP_k + FN_k}
\end{equation}

Using these, the F1-score for each class is:
\begin{equation}
F1_k = \frac{2 \cdot \text{Precision}_k \cdot \text{Recall}_k}{\text{Precision}_k + \text{Recall}_k}
\end{equation}

The macro F1-score is defined as the unweighted average across all classes:
\begin{equation}
\text{Macro-F1} = \frac{1}{K} \sum_{k=1}^{K} F1_k
\end{equation}

For completeness, we also report overall accuracy:
\begin{equation}
\text{Accuracy} = \frac{\sum_{k=1}^{K} TP_k}{N}
\end{equation}
where $N$ is the total number of samples.

In addition, we compute the weighted F1-score, which accounts for class imbalance by weighting each class-specific F1-score by its support $n_k$ (the number of samples in class $k$):
\begin{equation}
\text{Weighted-F1} = \frac{1}{N} \sum_{k=1}^{K} n_k \cdot F1_k
\end{equation}

While accuracy and weighted F1-score reflect overall performance dominated by frequent classes, the macro F1-score provides a balanced evaluation across all cell types and is therefore our primary metric.

\subsection{Spatial Transcriptomics Imputation}
\label{subsec:spatial-transcriptomics-imputation}

To evaluate performance on spatial transcriptomics imputation, we assess the agreement between imputed gene expression profiles and ground-truth measurements. Let $\mathbf{x}_i \in \mathbb{R}^G$ denote the ground-truth gene expression vector and $\hat{\mathbf{x}}_i \in \mathbb{R}^G$ the imputed vector for spot (or cell) $i$, where $G$ is the number of genes and $i \in \{1, \dots, N\}$ indexes spatial locations.

We employ two complementary metrics: \textbf{Pearson correlation} and \textbf{cosine similarity}, both computed at the spot level and averaged across all samples.

\paragraph{Pearson Correlation.}
Pearson correlation measures the linear relationship between imputed and ground-truth expression values. For each sample $i$, it is defined as:
\begin{equation}
r_i = 
\frac{
\sum_{g=1}^{G} \left(x_{ig} - \bar{x}_i\right)\left(\hat{x}_{ig} - \bar{\hat{x}}_i\right)
}{
\sqrt{\sum_{g=1}^{G} \left(x_{ig} - \bar{x}_i\right)^2}
\sqrt{\sum_{g=1}^{G} \left(\hat{x}_{ig} - \bar{\hat{x}}_i\right)^2}
}
\end{equation}
where $\bar{x}_i = \frac{1}{G}\sum_{g=1}^{G} x_{ig}$ and $\bar{\hat{x}}_i = \frac{1}{G}\sum_{g=1}^{G} \hat{x}_{ig}$ denote the mean expression values across genes for the ground-truth and imputed vectors, respectively.

The final Pearson correlation score is obtained by averaging across all samples:
\begin{equation}
\text{Pearson} = \frac{1}{N} \sum_{i=1}^{N} r_i
\end{equation}

\paragraph{Cosine Similarity.}
Cosine similarity measures the similarity in direction between imputed and ground-truth expression vectors, independent of their magnitudes. For each sample $i$, it is defined as:
\begin{equation}
\text{Cosine}_i =
\frac{
\sum_{g=1}^{G} x_{ig} \hat{x}_{ig}
}{
\sqrt{\sum_{g=1}^{G} x_{ig}^2} \;
\sqrt{\sum_{g=1}^{G} \hat{x}_{ig}^2}
}
\end{equation}

The overall cosine similarity is computed as:
\begin{equation}
\text{Cosine} = \frac{1}{N} \sum_{i=1}^{N} \text{Cosine}_i
\end{equation}

Pearson correlation captures linear agreement between imputed and true expression levels, while cosine similarity evaluates the consistency of expression patterns irrespective of scale. Together, these metrics provide a comprehensive assessment of imputation quality.

\subsection{Gene Perturbation Response Prediction}
\label{subsec:gene-perturbation-response}

To evaluate performance on gene perturbation response prediction, we measure the agreement between predicted and ground-truth \textit{differential gene expression} (DGE) profiles. Differential gene expression quantifies the change in gene expression induced by a perturbation. For each gene $g$ under perturbation $i$, it is defined as the difference between post-perturbation and pre-perturbation expression levels:
\begin{equation}
\Delta x_{ig} = x^{\text{post}}_{ig} - x^{\text{pre}}_{ig}
\end{equation}
where $x^{\text{pre}}_{ig}$ and $x^{\text{post}}_{ig}$ denote the expression of gene $g$ before and after the perturbation, respectively.

Let $\Delta \mathbf{x}_i \in \mathbb{R}^G$ denote the true differential expression vector and $\widehat{\Delta \mathbf{x}}_i \in \mathbb{R}^G$ the predicted differential expression vector for perturbation $i$, where $G$ is the number of genes and $i \in \{1, \dots, N\}$ indexes perturbation conditions.

We use the \textbf{Pearson correlation coefficient} to quantify similarity between predicted and true DGE profiles. For each perturbation $i$, the Pearson correlation is computed as:
\begin{equation}
r_i =
\frac{
\sum_{g=1}^{G} \left(\Delta x_{ig} - \overline{\Delta x}_i\right)\left(\widehat{\Delta x}_{ig} - \overline{\widehat{\Delta x}}_i\right)
}{
\sqrt{\sum_{g=1}^{G} \left(\Delta x_{ig} - \overline{\Delta x}_i\right)^2}
\sqrt{\sum_{g=1}^{G} \left(\widehat{\Delta x}_{ig} - \overline{\widehat{\Delta x}}_i\right)^2}
}
\end{equation}
where $\overline{\Delta x}_i = \frac{1}{G}\sum_{g=1}^{G} \Delta x_{ig}$ and $\overline{\widehat{\Delta x}}_i = \frac{1}{G}\sum_{g=1}^{G} \widehat{\Delta x}_{ig}$ denote the mean differential expression across genes for the ground-truth and predicted profiles, respectively.

The final evaluation metric is obtained by averaging across all perturbations:
\begin{equation}
\text{Pearson} = \frac{1}{N} \sum_{i=1}^{N} r_i
\end{equation}

Pearson correlation captures the linear agreement between predicted and true differential expression patterns across genes, providing a robust measure of how well the model recovers the direction and relative magnitude of gene-level responses to perturbations.

\section{Cell Ontologies and Type Distributions}
\label{cell-type-dist}

\begin{figure}[h]
  \centering
  \includegraphics[width=1\linewidth]{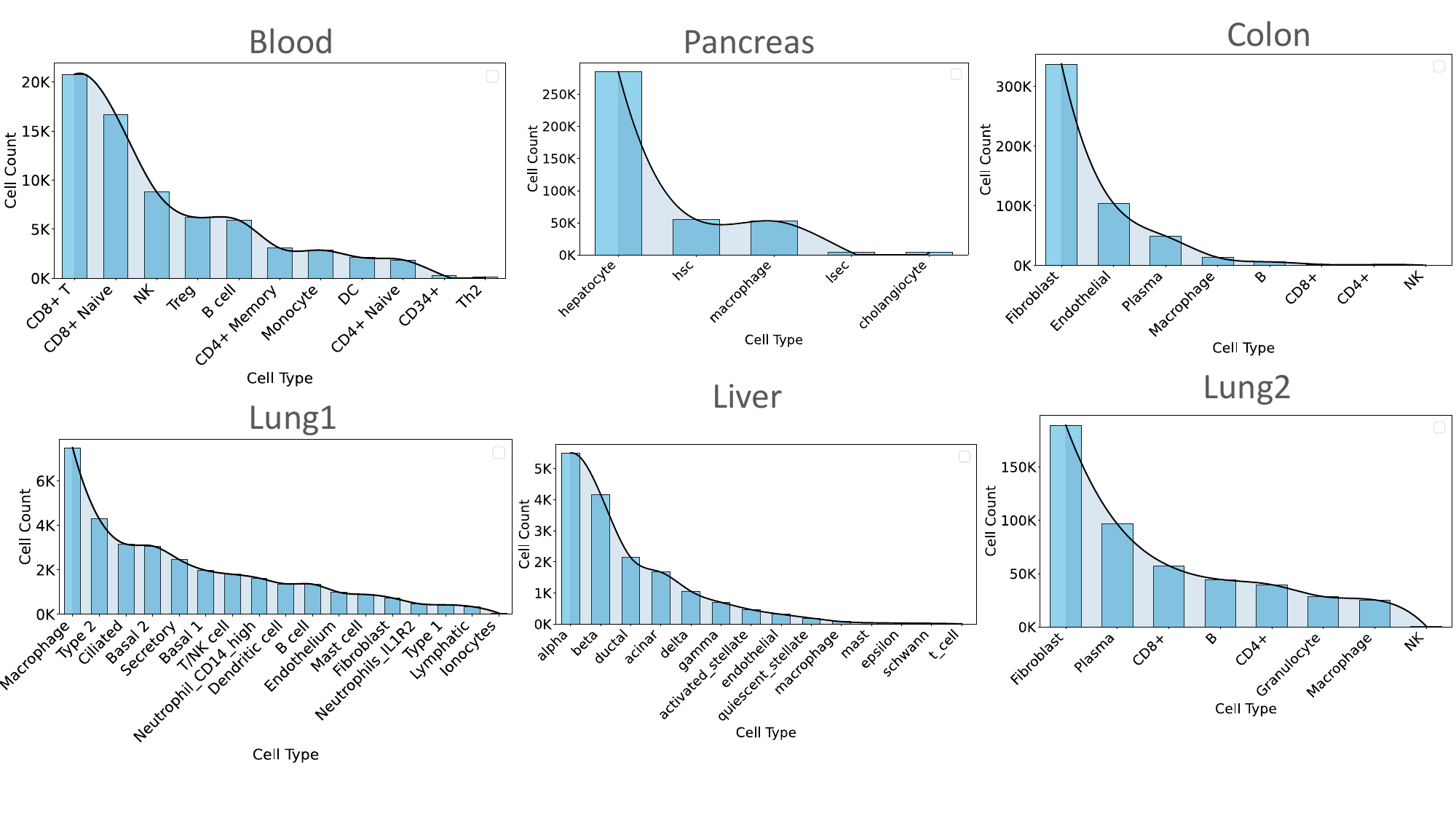}
  \caption{Long-tailed distributions of single-cell datasets}
  \label{fig:long-tail-appendix}
\end{figure}

Figure~\ref{fig:long-tail-appendix} illustrates the long-tail distributions of cell types in many of the single-cell datasets. 

\section{Long tail quantification}
\label{long-tail-analysis}

To quantify the degree of class imbalance across datasets, we estimated 
\textit{power-law tail exponents} ($\alpha$) from the distribution of cell 
type frequencies. Specifically, we fit a log--log regression to the 
\textit{complementary cumulative distribution function (CCDF)} of the rarest 
30\% of categories, following common practice in heavy-tailed data analysis. 
The CCDF of a power-law distributed variable $X$ is given by
\[
P(X \geq x) \;\sim\; x^{-\alpha}
\]
where smaller values of $\alpha$ correspond to heavier tails, reflecting 
stronger dominance of a few frequent types and a larger fraction of rare ones. 
As summarized in Table~\ref{tab:tail_exponents}, the fitted exponents range 
from $\alpha \approx 0.2$ (Custom3) to $\alpha \approx 0.6$ (MS), confirming 
that single-cell datasets exhibit highly skewed and heavy-tailed distributions. 
For the Liver dataset, the number of categories ($n=5$) was too small to 
reliably estimate a tail exponent. These results provide a quantitative basis 
for treating single-cell learning tasks as inherently long-tailed 
recognition problems, motivating the need for methods that explicitly address 
rare classes.

\begin{table}[h]
\centering

\begin{tabular}{lcccc}
\hline
\textbf{Dataset} & \textbf{\# Cell Types} & \textbf{Estimated Tail Exponent ($\alpha$)} & \textbf{Fit $R^2$} \\
\hline
Pancreas   & 17 & 0.45  & 0.583 \\
MS          & 14 & 0.631 & 0.693 \\
Blood    & 11 & 0.37  & 1.000 \\
\hline
\end{tabular}
\caption{Estimated tail exponents for cell type distributions across datasets. Smaller $\alpha$ values indicate heavier tails.}
\label{tab:tail_exponents}
\end{table}

\section{Additional dataset details}
\label{sec:additional-dataset-details}

\begin{table*}[t]
\centering
\small
\setlength{\tabcolsep}{6pt}
\renewcommand{\arraystretch}{1.05}

\resizebox{\textwidth}{!}{%
\begin{tabular}{llllc}
\toprule
\textbf{Dataset}
& \textbf{Number of sources}
& \textbf{Downstream task}
& \textbf{Cell type}
& \textbf{Count} \\
\midrule

\multirow[t]{5}{*}{LiverST1}
& \multirow[t]{5}{*}{1}
& \multirow[t]{5}{*}{\makecell[l]{Cell identity prediction \\ Spatial transcriptomics imputation}}
& hepatocyte & 228,654 \\
& & & hsc & 41,043 \\
& & & macrophage & 34,861 \\
& & & cholangiocyte & 3,110 \\
& & & lsec & 2,140 \\
\midrule

\multirow[t]{5}{*}{LiverST2}
& \multirow[t]{5}{*}{1}
& \multirow[t]{5}{*}{\makecell[l]{Cell identity prediction \\ Spatial transcriptomics imputation}}
& hepatocyte & 56,286 \\
& & & macrophage & 17,808 \\
& & & hsc & 13,992 \\
& & & lsec & 1,690 \\
& & & cholangiocyte & 632 \\
\midrule

\multirow[t]{11}{*}{Blood}
& \multirow[t]{11}{*}{1}
& \multirow[t]{11}{*}{Cell identity prediction}
& CD8+ Cytotoxic T & 20,757 \\
& & & CD8+/CD45RA+ Naive Cytotoxic & 16,645 \\
& & & CD56+ NK & 8,775 \\
& & & CD4+/CD25 T Reg & 6,185 \\
& & & CD19+ B & 5,877 \\
& & & CD4+/CD45RO+ Memory & 3,059 \\
& & & CD14+ Monocyte & 2,847 \\
& & & Dendritic & 2,095 \\
& & & CD4+/CD45RA+/CD25$^{-}$ Naive T & 1,871 \\
& & & CD34+ & 242 \\
& & & CD4+ T Helper2 & 97 \\
\midrule

\multirow[t]{14}{*}{Pancreas}
& \multirow[t]{14}{*}{5}
& \multirow[t]{14}{*}{Cell identity prediction}
& alpha & 5,147 \\
& & & beta & 3,972 \\
& & & ductal & 1,704 \\
& & & acinar & 1,353 \\
& & & delta & 981 \\
& & & PP & 638 \\
& & & PSC & 597 \\
& & & endothelial & 289 \\
& & & macrophage & 52 \\
& & & mast & 32 \\
& & & epsilon & 28 \\
& & & schwann & 13 \\
& & & t\_cell & 7 \\
& & & MHC class II & 5 \\

\midrule

\multirow[t]{14}{*}{PBMC10K}
& \multirow[t]{14}{*}{1}
& \multirow[t]{14}{*}{Cell identity prediction}
& CD14+ Monocytes & 2,992 \\
& & & CD4 Memory & 1,596 \\
& & & CD4 Naive & 1,047 \\
& & & pre-B cell & 959 \\
& & & Double negative T cell & 592 \\
& & & B cell progenitor & 460 \\
& & & NK dim & 435 \\
& & & CD8 effector & 383 \\
& & & CD8 Naive & 337 \\
& & & CD16+ Monocytes & 328 \\
& & & NK bright & 109 \\
& & & Dendritic cell & 74 \\
& & & pDC & 68 \\
& & & Megakaryocytes & 52 \\

\midrule
\multirow[t]{10}{*}{Heart}
& \multirow[t]{10}{*}{1}
& \multirow[t]{10}{*}{Cell identity prediction}
& Cardiomyocyte & 91,957 \\
& & & Fibroblast & 83,220 \\
& & & Endothelial & 27,923 \\
& & & Pericyte & 18,467 \\
& & & Macrophage & 17,468 \\
& & & Adipocyte & 8,658 \\
& & & Vascular smooth muscle & 5,740 \\
& & & Neuronal & 1,568 \\
& & & Lymphocyte & 1,503 \\
& & & Cytoplasmic cardiomyocyte & 30,765 \\
    
\bottomrule
\end{tabular}%
}
\caption{Dataset composition and downstream tasks across the evaluated benchmarks (1/2).}
\label{tab:dataset_composition}
\end{table*}

\begin{table*}[t]
\centering
\small
\setlength{\tabcolsep}{6pt}
\renewcommand{\arraystretch}{1.05}

\resizebox{\textwidth}{!}{%
\begin{tabular}{llllc}
\toprule
\textbf{Dataset}
& \textbf{Number of sources}
& \textbf{Downstream task}
& \textbf{Cell type}
& \textbf{Count} \\
\midrule

\multirow[t]{10}{*}{Liver}
& \multirow[t]{10}{*}{2}
& \multirow[t]{10}{*}{Cell identity prediction}
& Hepatocytes & 6,587 \\
& & & NK, NKT and T cells & 5,084 \\
& & & LSEC & 1,994 \\
& & & macrophage & 1,192 \\
& & & cholangiocytes & 1,141 \\
& & & MSEC & 566 \\
& & & Plasma cells & 511 \\
& & & B cells & 258 \\
& & & Erythroid cells & 93 \\
& & & Stellate and myofibroblasts & 65 \\
\midrule

\multirow[t]{23}{*}{Myeloid}
& \multirow[t]{23}{*}{9}
& \multirow[t]{23}{*}{Cell identity prediction}
& Mono CD16 & 11,214 \\
& & & Mono CD14 & 10,546 \\
& & & Macro C1QC & 7,927 \\
& & & cDC2 CD1C & 7,883 \\
& & & Macro NLRP3 & 3,903 \\
& & & Mast KIT & 3,868 \\
& & & Macro LYVE1 & 3,419 \\
& & & Macro SPP1 & 3,279 \\
& & & Macro IL1B & 2,363 \\
& & & Macro GPNMB & 2,264 \\
& & & Macro INHBA & 2,047 \\
& & & cDC2 CXCR4hi & 1,851 \\
& & & Mono CD14CD16 & 1,608 \\
& & & Macro FN1 & 1,319 \\
& & & cDC2 CD1A & 1,209 \\
& & & cDC2 IL1B & 1,208 \\
& & & pDC LILRA4 & 786 \\
& & & cDC3 LAMP3 & 740 \\
& & & Macro ISG15 & 726 \\
& & & cDC2 FCN1 & 720 \\
& & & cDC1 CLEC9A & 673 \\
& & & cDC2 ISG15 & 302 \\
& & & cDC2 CXCL9 & 171 \\

\midrule

\multirow[t]{8}{*}{Blood(Kang)}
& \multirow[t]{8}{*}{1}
& \multirow[t]{8}{*}{Gene perturbation response prediction}
& CD4 T cells & 11,238 \\
& & & CD14+ Monocytes & 5,697 \\
& & & B cells & 2,651 \\
& & & NK cells & 1,716 \\
& & & CD8 T cells & 1,621 \\
& & & FCGR3A+ Monocytes & 1,089 \\
& & & Dendritic cells & 529 \\
& & & Megakaryocytes & 132 \\

\midrule

\multirow[t]{18}{*}{MS}
& \multirow[t]{18}{*}{1}
& \multirow[t]{18}{*}{Cell identity prediction}
& cortical layer 2-3 excitatory neuron B & 3,197 \\
& & & cortical layer 4 excitatory neuron & 2,788 \\
& & & cortical layer 5-6 excitatory neuron & 2,729 \\
& & & cortical layer 2-3 excitatory neuron A & 2,324 \\
& & & VIP-expressing interneuron & 1,820 \\
& & & astrocyte & 1,484 \\
& & & PVALB-expressing interneuron & 1,447 \\
& & & oligodendrocyte A & 1,269 \\
& & & pyramidal neuron & 941 \\
& & & oligodendrocyte precursor cell & 921 \\
& & & mixed glial cell & 633 \\
& & & SV2C-expressing interneuron & 578 \\
& & & SST-expressing interneuron & 430 \\
& & & mixed excitatory neuron & 299 \\
& & & endothelial cell & 208 \\
& & & microglial cell & 122 \\
& & & phagocyte & 85 \\
& & & oligodendrocyte C & 37 \\

\midrule
\multirow[t]{10}{*}{Lung}
& \multirow[t]{10}{*}{1}
& \multirow[t]{10}{*}{Cell identity prediction}
& Malignant & 10,538 \\
& & & T cell & 6,721 \\
& & & Macrophage & 6,487 \\
& & & B cell & 2,177 \\
& & & Fibroblast & 1,583 \\
& & & Epithelial & 944 \\
& & & Mast & 764 \\
& & & NK cell & 639 \\
& & & Endothelial & 492 \\
& & & Dendritic & 127 \\

\bottomrule
\end{tabular}%
}
\caption{Dataset composition and downstream tasks across the evaluated benchmarks (2/2).}
\label{tab:dataset_composition}
\end{table*}

\end{document}